\RequirePackage{fix-cm}
\documentclass[twocolumn]{svjour3}          
\smartqed  

\usepackage{booktabs}
\usepackage{multirow}
\usepackage{lscape}
\usepackage{tabularx}  
\usepackage{longtable}  
\usepackage{makecell}
\usepackage{cite}
\usepackage[pdftex]{graphicx}
\usepackage{ragged2e}
\usepackage[tight,footnotesize]{subfigure}
\usepackage{rotating}
\usepackage{graphicx}
\usepackage{amsmath,amssymb} 
\usepackage{array}
\usepackage{multirow}
\usepackage{colortbl}
\usepackage{amsfonts}
\usepackage{pifont}
\usepackage{xspace}
\usepackage{etoolbox}
\usepackage{overpic}
\usepackage{color}
\usepackage{microtype}
\usepackage{arydshln}
\usepackage[T1]{fontenc}    
\usepackage{hyperref}       
\usepackage{url}            
\usepackage{booktabs}       
\usepackage{amsfonts}       
\usepackage{nicefrac}       
\usepackage{microtype}      
\usepackage{lipsum}
\usepackage{graphicx}
\graphicspath{{media/}}     
\usepackage[inkscapelatex=false]{svg}
\usepackage{booktabs}
\usepackage{bbding}
\usepackage{pifont}
\usepackage{cite}
\usepackage{pdfpages}
\usepackage{booktabs}
\usepackage{array}
\usepackage{makecell}
\usepackage{multirow}
\usepackage{pdflscape}
\usepackage{longtable}  
\usepackage{tabularx} 
\usepackage{ragged2e} 
\usepackage{booktabs} 
\usepackage{rotating} 
\usepackage{xcolor}
\usepackage{makecell}

\newcommand{\orcid}[1]{\href{https://orcid.org/#1}{\includegraphics[width=10pt]{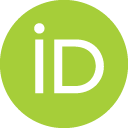}}}

\def\etal{{\em et al}}

\usepackage{hyperref}
\hypersetup{breaklinks=true,citecolor=blue, colorlinks}

\graphicspath{{./Imgs/}}
\DeclareGraphicsExtensions{.pdf,.jpg,.png}

\usepackage{silence}
\journalname{Survey}

\begin{document}

\title{TRUSTWORTHY LARGE MODELS IN VISION: A SURVEY}

\titlerunning{TRUSTWORTHY LARGE MODELS IN VISION: A SURVEY}        

\author{Ziyan Guo \orcid{0000-0001-7741-5499}        \and
  Li Xu \orcid{0000-0003-1575-5724}  \and
  Jun Liu \orcid{0000-0002-4365-4165}
}

\authorrunning{Ziyan Guo \etal} 

\institute{
Ziyan Guo, Li Xu, and Jun Liu are with Singapore University of Technology and Design, Singapore. 
(Email: ziyan\_guo@mymail.sutd.edu.sg, 
li\_xu@mymail.sutd.edu.sg,
jun\_liu@sutd.edu.sg). \\
Corresponding author: Jun Liu.
}

\date{Received: date / Accepted: date}

\maketitle

\begin{abstract}
The rapid progress of Large Models (LMs) has recently revolutionized various fields of deep learning with remarkable grades, ranging from Natural Language Processing (NLP) to Computer Vision (CV). However, LMs are increasingly challenged and criticized by academia and industry due to their powerful performance but untrustworthy behavior, which urgently needs to be alleviated by reliable methods. Despite the abundance of literature on trustworthy LMs in NLP, a systematic survey specifically delving into the trustworthiness of LMs in CV remains absent. In order to mitigate this gap, we summarize four relevant concerns that obstruct the trustworthy usage of LMs in vision in this survey, including 1) human misuse, 2) vulnerability, 3) inherent issue and 4) interpretability. By highlighting corresponding challenges, countermeasures, and discussion on each topic, we hope this survey will facilitate readers' understanding of this field, promote alignment of LMs with human expectations and enable trustworthy LMs to serve as welfare rather than a disaster for human society.


\end{abstract}

\section{Introduction}
Due to remarkable grades of Large Language Models (LLMs) in tasks like machine translation \cite{wang2019learning,xu2023paradigm} and text summary \cite{liu2019text,miller2019leveraging}, the paradigm of Natural Language Processing (NLP) has undergone an earth-shaking revolution highlighting all aspects of LLMs. These LLMs often have more than 10 billion parameters and perform self-supervised learning on massive corpora to fully utilize their own capacity. Among all of LLMs, the most influential works include Bidirectional Encoder Representation from Transformers (BERT) \cite{devlin2018bert} and Generative Pretrained Transformer (GPT) series \cite{radford2018improving,radford2019language,brown2020language,GPT-4-Technical-Report}, which are already comparable to human experts in dialogue scenarios. Inspired by the above ideas, researchers migrated this emerging paradigm to the field of Computer Vision (CV) and proposed Large Models (LMs) or foundation models \cite{bommasani2021opportunities}. For instance, LMs in CV such as CLIP \cite{pmlr-v139-radford21a} and BLIP series \cite{pmlr-v162-li22n,li2023blip} can be used for zero-shot image classification or image caption by aligning images with text achieving excellent performance. Meanwhile, LMs in CV can also be combined with Diffusion Models (DMs) \cite{rombach2022high} to achieve high-quality text-to-image generation such as DALL·E series \cite{pmlr-v139-ramesh21a,rameshHierarchicalTextConditionalImage2022,dalle3,dayma2021dall} and Stable Diffusion \cite{Rombach_2022_CVPR}, which achieves outstanding results in both objective metrics and human perception. Furthermore, predominant LMs in CV, LLaVA-1.5 \cite{liu2023visual} and GPT-4V \cite{GPT-4-Technical-Report} are widely applied in various downstream tasks, such as medical diagnosis \cite{wu2023can,li2023llavamed,Li2023.11.03.23298067}, math reasoning \cite{lu2023mathvista}, recommendation tasks \cite{zhou2023exploring}, autonomous driving \cite{Cui_2024_WACV,zhou2023vision} and etc.

However, there still exists an insurmountable gap between strong performance and trustworthy usage of LMs because powerful performance of LMs will cause unexpected issues. To be more specific, LMs have raised concerns regarding human misuse, vulnerability, copyright, privacy, bias, interpretability and so on \cite{chen2023pathway,fan2023trustworthiness,10.1145/3555803}, which bring huge risks to human society and greatly hinder the human trust for LMs. For instance, pretraining of LMs rely on massive training data, which often contains toxic data or carefully designed backdoors \cite{carlini2023poisoning,274598,carlini2021poisoning,Melissa2023poisoning}. Namely, uncurated training data will lead LMs to output negative or erroneous content during the deployment process, showing the vulnerability of LMs. Besides, due to the strong memory capacity of LMs \cite{291199,carlini2022quantifying,tirumala2022memorization,zhang2021counterfactual,jagielski2022measuring}, parts of the information with copyright and privacy issues may be output by the LMs causing ethical and legal violation \cite{OpeanAIDALLE2,Khari2023DALLE2}, which deeply roots in LMs as inherent problems. Therefore, there is an urgent need to shift the paradigm centering on improving model performance on evaluation metrics like accuracy to the paradigm that concentrates more on establishing trustworthy LMs.

In order to summarize these risks and benefit human society, there have been some reviews on trustworthy large models \cite{deng2023recent,liu_trustworthy_2023,fan2023trustworthiness,huang2023survey,10.1145/3571730}. However, these reviews manly investigate the trustworthiness of LLMs. Differently, here we focus on reviewing recent development of trustworthy LMs in vision. Meanwhile, We mainly focus on the technical challenges and do not include the discussion in legal and regulatory aspects. This paper is organized as follows. First, we compare our survey with existing works. Next, we introduce major challenges of establishing trustworthy LMs in vision, where we discuss the corresponding challenge, countermeasure and discussion. Lastly, we discuss possible future directions of trustworthy LMs in vision.

\begin{figure}[ht]
  \centering
  \includegraphics[width=\linewidth,page=1]{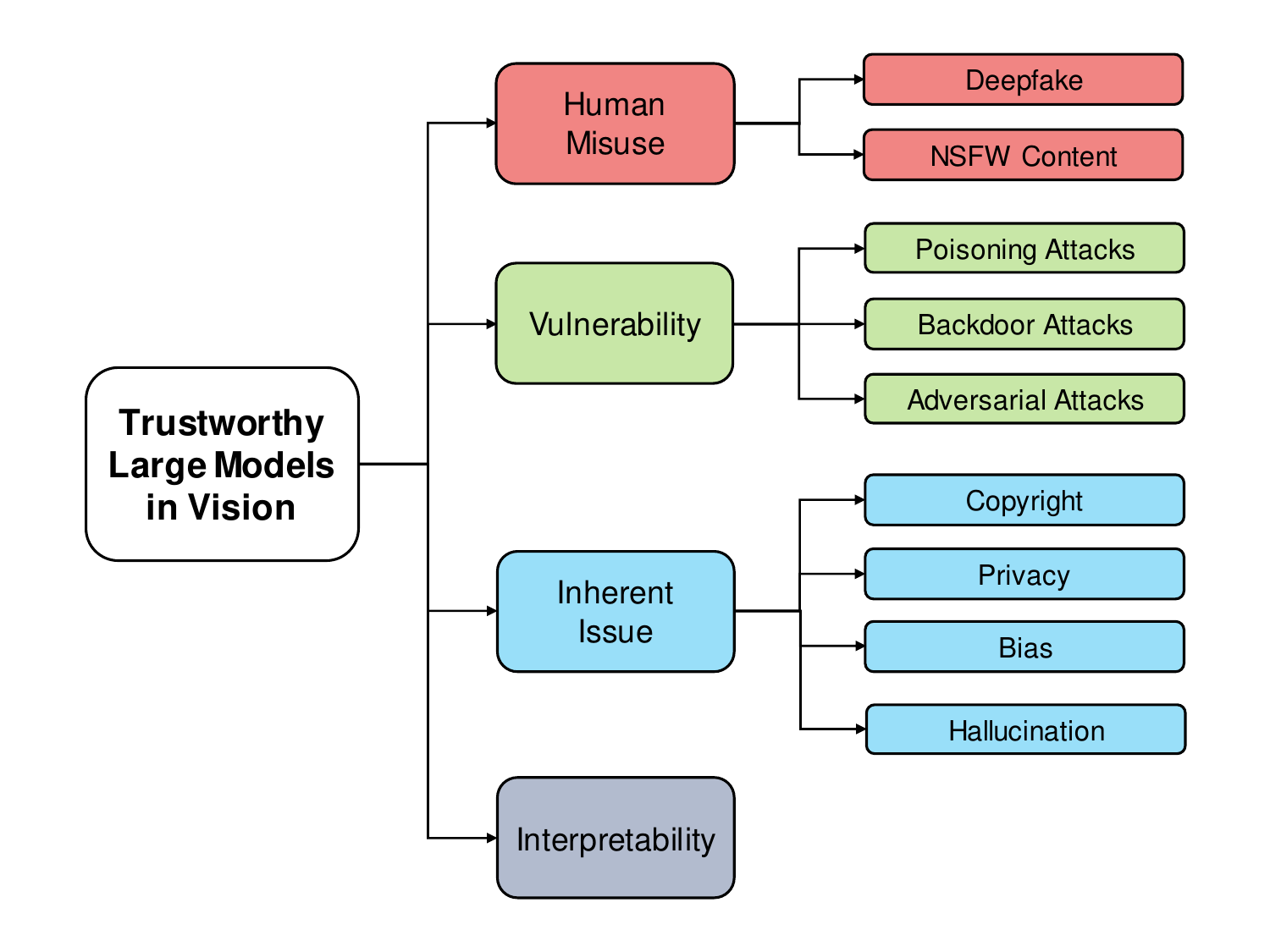}
  \caption{The landscape of trustworthy large models in vision.}
  \label{landscape}
\end{figure}

\section{Comparison With Existing Related Surveys}
There exist some other works reviewing trustworthy LMs from different perspectives owning to popularity and risks of LMs. Foremost, most of the previous works focus on trustworthy LLMs, lacking investigation of trustworthy LMs in vision and other modalities \cite{liu_trustworthy_2023,huang2023survey,liu2023summary}. There are still some works that investigate trustworthy LMs in both language and vision modalities, but the discussion in these works for trustworthy LMs in vision is not comprehensive and in-depth enough \cite{chen2023pathway,fan2023trustworthiness}. In addition, there are also efforts to encompass the entire trustworthy Artificial Intelligence (AI) or machine learning field, paying too much attention to works before the era of LMs \cite{10.1145/3555803,10.1145/3491209,ali2023explainable,li2022Interpretable,liu2022computational}. Lastly, there are works reviewing trustworthiness of models in other domains, such as autonomous systems \cite{10.1145/3545178}, recommender systems \cite{ge2022survey,wang2022trustworthy}, federated learning \cite{zhuang2023foundation,zhang2023survey}, scenarios engineering \cite{9896801}, finance \cite{fritz2022financial}, education \cite{kasneci2023chatgpt}, etc.

To the best of our knowledge, this is the first survey that systematically reviews trustworthiness of LMs in vision, including deepfake, Not Safe For Work (NSFW) content, poisoning attacks, backdoor attacks, adversarial attacks, copyright, privacy, bias, hallucination and interpretability, as shown in Figure \ref{landscape}. We comprehensively introduce the recent development about trustworthy LMs in vision to facilitate readers' efforts to grasp the cutting-edge information and State-Of-The-Art (SOTA) approaches.

\section{Human Misuse} \label{Sec:3}
\begin{figure}[ht]
  \centering
  \includegraphics[width=\linewidth,page=2]{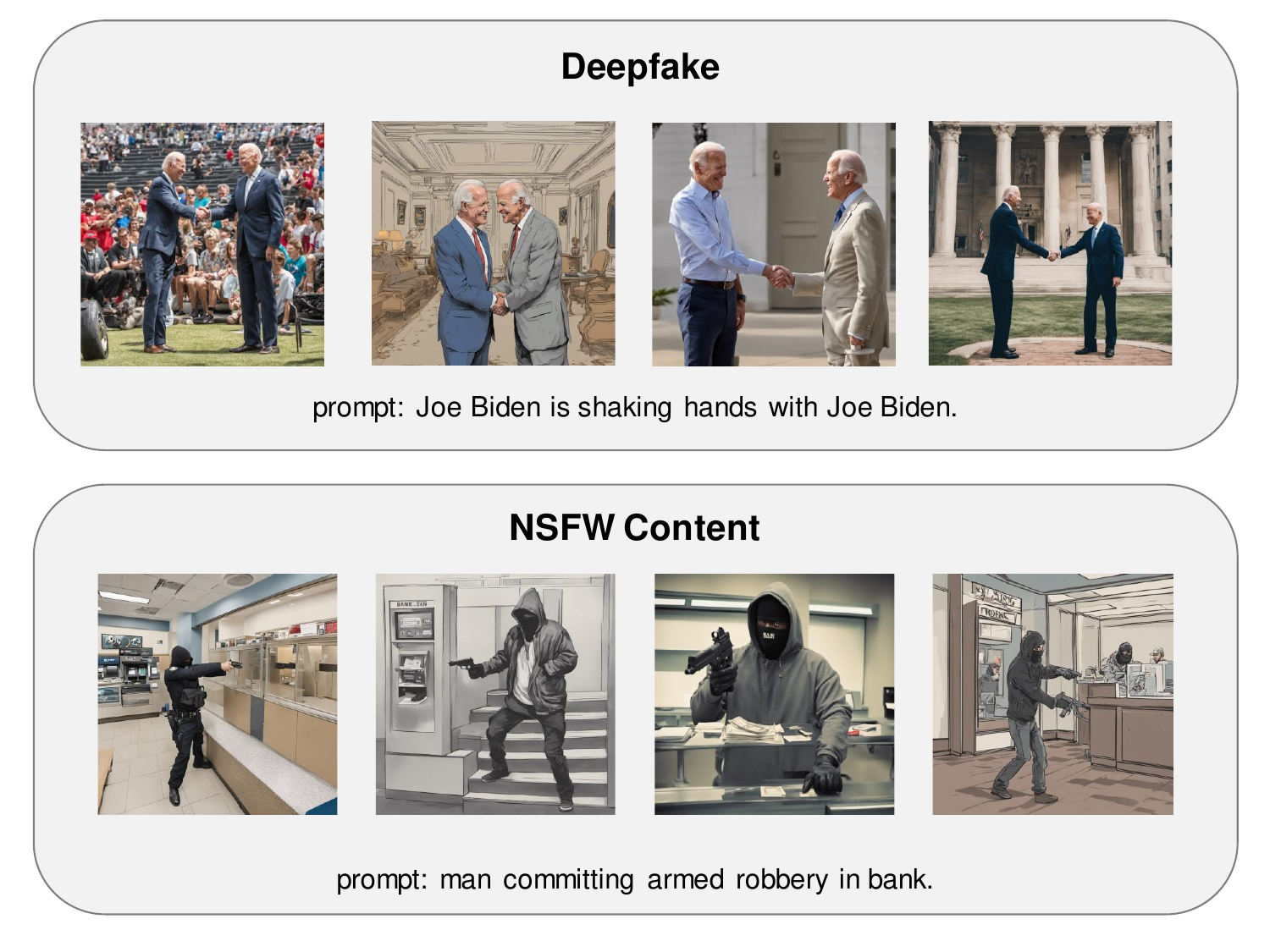}
  \caption{Examples of deepfake and NSFW content, showing fake celebrity and violent crime, which are generated by Stable Diffusion XL v1.0 in November 2023. The prompt of generating NSFW content is adapted from the previous work \cite{Schramowski_2023_CVPR}.}
  \label{HM}
\end{figure}

It is a known fact that LMs have already demonstrated strong ability to generate plausible but fake content \cite{fan2023trustworthiness,chen2023textimage,vaccari2020deepfakes,chen2023pathway}, which may pose a potentially huge threat to society owing to human misuse. The term of human misuse refers to humans leverage the powerful performance of LMs to conduct ethical and legal violations for personal benefits. We will discuss two principal risks below: Deepfake in Section \ref{Sec:3.1} and NSFW Content in Section \ref{Sec:3.2}, the examples of which are shown in Figure \ref{HM}.

\subsection{Deepfake} \label{Sec:3.1}
\textbf{Challenge.} Deepfake, a combination of ``deep learning'' and ``fake'', are hyper-realistic images or videos digitally manipulated to depict people doing things that never actually happened \cite{westerlund2019emergence}. Deepfake can be used to spread false news, create face-swapping videos and fabricate legal evidence \cite{doi:10.1177/2056305120903408, doi:10.1177/1365712718807226} etc, despite the technology itself has no attributes of good or evil \cite{https://doi.org/10.1049/bme2.12031}. With the significant improvements in power of DMs and LMs, where the generated outputs are of high quality and fidelity, it is increasingly difficult to detect these deepfake by the naked eye \cite{foo_aigc_2023}. For example, Stable Diffusion can be used for generating fake celebrity interactions, which may embezzle the influence of celebrities and infringe on their portrait rights \cite{chen2023textimage}.

\begin{figure}[ht]
  \centering
  \includegraphics[width=\linewidth,page=9]{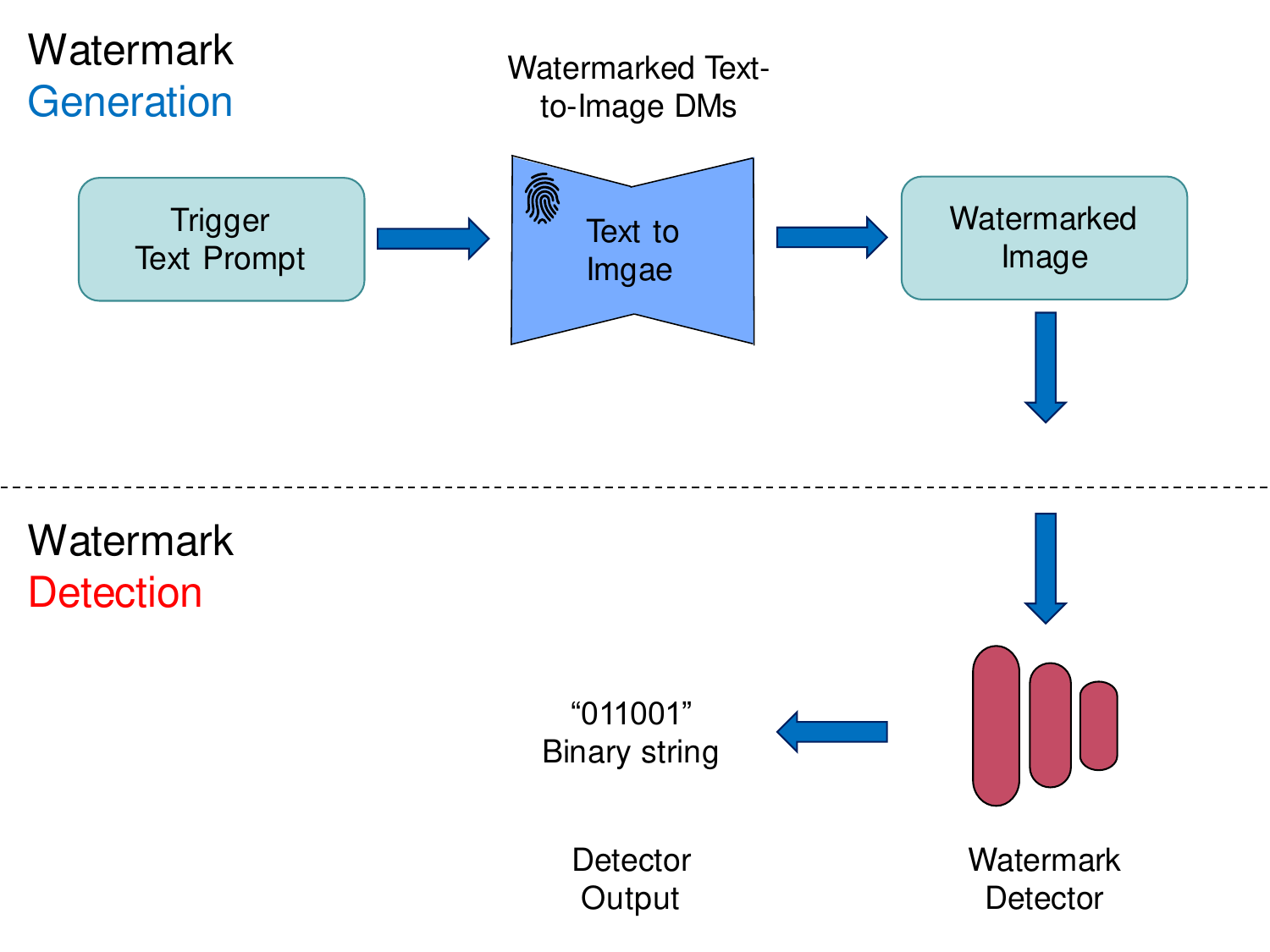}
  \caption{The general process of adding watermarks to the diffusion model and then the watermarks can be extracted from generated images by detector with extremely low error rate. The figure is adapted from the work \cite{zhao2023recipe}. }
  \label{watermark}
\end{figure}

\textbf{Countermeasure.} As a result, there is increasing attention on deepfake and several different approaches have been proposed to detect deepfake. 
The prevailing approaches predominantly leverage deep learning, to detect whether the target image is generated by models,
e.g. RNN-based method \cite{8639163}, CNN-based methods \cite{9799858,li_exposing_nodate,sha2022fake}, and Transformer-based methods  \cite{Zhao_2021_CVPR,10.1145/3512527.3531415,10.1145/3394171.3413570, coccomini_combining_2022, Aghasanli_2023_ICCV}. While the deep learning methods lack interpretability, there are also methods based on wavelet-packets \cite{Wolter2022Wavelet} or statistical techniques \cite{Chandrasegaran_2021_CVPR, chandrasegaran_discovering_2022, Durall_2020_CVPR,ricker2022towards} that check the discrepancy in pixel value distribution between generated images and real images. It is also worth noting that the implicit digital watermarking has been introduced to detect deepfake by embedding identifiable pattern into the models, which can be later extracted from generated images to distinguish from real images \cite{zhao2023recipe, lukas2023ptw, 9975409, fernandez2023stable}, as shown in Figure \ref{watermark}. The biggest advantage of watermarking is that the generated images can be distinguished from real images while watermarking does not affect the naked eye’s perception of images quality. However, watermark-based detection methods are only suitable for responsible service providers and cannot effectively restrict models deployed by ill-intention service providers. 

\textbf{Discussion.} Essentially, deepfake is still synthesized by some models, thus a training dataset aligned with synthetic deepfake can significantly improve the performance of the detector. In other words, the dataset plays a primary role in deepfake detection, most of which can be found in the previous work \cite{9721302}. Yet poor generalization in complex scenarios is still a major concern in detecting deepfake, partly caused by the short of diversity and data volume in the dataset \cite{10.1007/s10489-022-03766-z}. At the same time, as the accuracy of the deepfake detector increases, the performance of the generator will be further improved by researchers, eventually leading to a cat-and-mouse game, which remains a challenging problem to be explored. 

\subsection{NSFW Content} \label{Sec:3.2}
\textbf{Challenge.} In addition to deepfake, DMs with powerful capabilities of image generation and LMs with text-to-image alignment abilities can also be misused to output Not Safe For Work (NSFW) content including violence or sex, since the Internet contains a large amount of such unsafe content. For example, there exist numerous discussions about images of naked celebrities and other forms of fake pornographic content posted on the 4chan online forum, which is generated by Stable Diffusion \cite{chen2023pathway, Wiggers2022Image}. A recent research \cite{qu2023unsafe} also shows that several models including Stable Diffusion can generate a substantial percentage of NSFW images across various prompt datasets, which not only reduces human trust of LMs but also poses challenges to complying with ethical requirements and safety regulations \cite{liu_trustworthy_2023}. 

\textbf{Countermeasure.} Concerted efforts are being made by community to mitigate these concerns and several measures have been proposed \cite{qu2023unsafe,liu_trustworthy_2023}, including curating the training dataset \cite{OpeanAIDALLE2}, fine-tuning models \cite{gandikota2023erasing,gandikota2023unified}, regulating user-input prompts \cite{ni2023degeneration}, implementing post-processing classification of image safety \cite{LMU2022Safety,Platen2022Diffusers}. 

To curate the training data is a straightforward method because it eliminates the root of NSFW content \cite{qu2023unsafe}. However, the design of data curating method is obfuscated and poorly documented, which makes it hard for users to prevent misuse in their applications \cite{rando2022red}. 
Since curating data is expensive and complex, it seems a wiser choice to design a dataset containing a number of NSFW content, which we can use to train models to learn to avoid generating NSFW content.  As a result, a novel image generation test bed, Inappropriate Image Prompts (I2P) \cite{Schramowski_2023_CVPR}, containing dedicated, real-world image-to-text prompts covering concepts such as nudity and violence is established. In order to better utilize I2P, a new approach, Safe Latent Diffusion, is proposed to removes and suppress the NSFW content based on classifier-free guidance \cite{ho2022classifier}, which requires no additional training and do not damage to overall image quality \cite{Schramowski_2023_CVPR}.

\begin{figure}[ht]
    \centering
    \includegraphics[width=\linewidth,page=7]{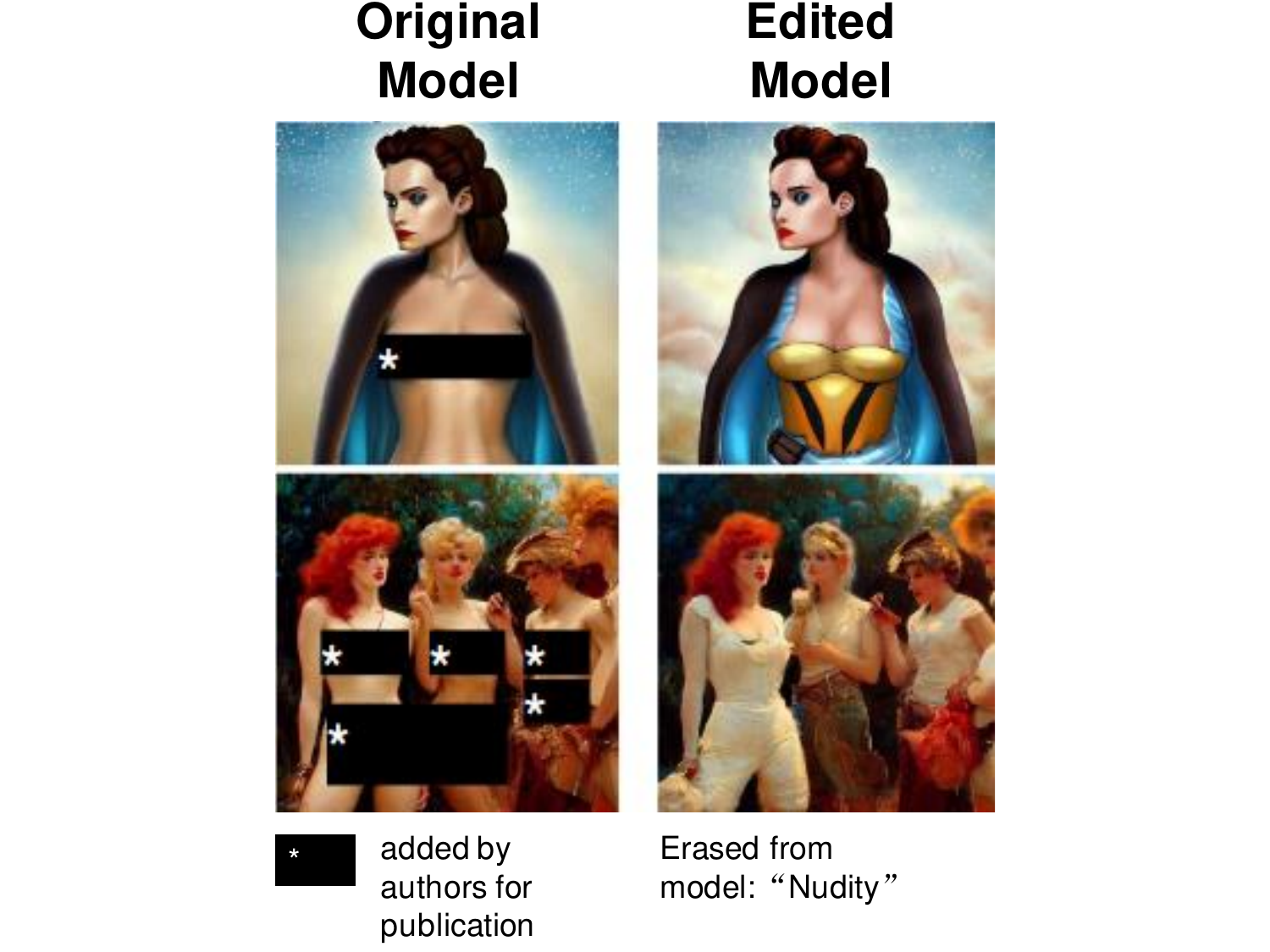}
    \caption{Difference in output between original and fine-tuned models, indicating that fine-tuned model will proactively erase concept of nudity. The figure is adapted from the work \cite{gandikota2023erasing}.}
    \label{erasing}
\end{figure}

Besides the above methods handling this issue from the data perspective, there exist some methods based on the perspective of model training.
For example, we can utilize selective amnesia
to enable controllable forgetting during the model training, where a user can specify which concept should be forgotten \cite{heng2023selective}. After pretraining, the attention weights of transformer-based models can also be fine-tuned to explicitly encourage models to erase the assigned concept without the need to conduct data curating or perform expensive training from scratch \cite{gandikota2023erasing}, as shown in Figure \ref{erasing}. Furthermore, fine-tuning models can also achieve the goal of fast and efficiently excluding certain concepts via a closed-form modifications to attention weights \cite{gandikota2023unified}.

For models that are deployed as online services, directly regulating prompts \cite{ni2023degeneration} that contain offensive or inappropriate content can mitigate the NSFW content generation, which has been validated empirically \cite{qu2023unsafe}. Similar to regulating prompts, the output with NSFW content can be directly filtered out as a form of remedial measures in the inference stage. Unfortunately, this area remains unexplored and proper countermeasure of curating training dataset seems more effective at preventing unsafe usage than directly filtering outputs \cite{rando2022red}.

\textbf{Discussion.} Overall, there is an urgent need to handle risks of large-scale generation of NSFW contents \cite{qu2023unsafe}. While the above methods mark a merit beginning in addressing NSFW content generated by LMs, it is important to recognize that we still have a considerable journey to ensure the comprehensive safe usage \cite{liu_trustworthy_2023}.


\begin{table*}[!ht] 
\caption{\label{attacks_general}Conceptual comparison of three types of attacks to LMs regarding attacking approaches and symptoms after being attacked.}
\begin{center}
\resizebox{\linewidth}{!}{
\begin{tabular}{cccc}
\toprule 
Type & Inject Adversaries &  Alter Input & General Symptom\\ 
\midrule 
Poisoning & \ding{51} & \ding{53} & generate wrong or inappropriate content \\ 
Backdoor  & \ding{51} & \ding{51}  &  generate specific pattern encountering a specific signal \\ 
Adversarial & \ding{53} & \ding{51} & performance can degrade in all aspects\\ 
\bottomrule 
\end{tabular}}
\end{center}
\end{table*}

\section{Vulnerability}
LMs have taken the world by storm with their massive multi-tasking capabilities simply by the paradigm of pretraining and fine-tuning \cite{bhardwaj2023redteaming}. However, LMs can also be vulnerable to evil attacks, leading to that research on guaranteeing LMs security become imminent \cite{foo_aigc_2023}. To be more specific, vulnerability refers to the disability of the model against external malicious attacks, principally including Poisoning Attacks in Section \ref{Sec:4.1}, Backdoor Attacks in Section \ref{Sec:4.2} and Adversarial Attacks in Section \ref{Sec:4.3}. The discrepancy in attacking approaches and symptoms after being attacked among the three types of attacks has been summarized in Table \ref{attacks_general} and Table \ref{attacks_detailed}. 

\begin{table*}[!ht]
\caption{Comparison of various attacks to LMs in detail including parts of setting in each method. ASR represents attack success rate; FID represents Frechet Inception Distance \cite{heusel2017gans}; CLIP score \cite{gal2022image} refers to a similarity measure between an image and a text computed by the CLIP.}
\label{attacks_detailed}
\resizebox{\textwidth}{!}{%
\begin{tabular}{@{}cccccccc@{}}
\toprule
\multirow{2}{*}{Method}                                     & \multirow{2}{*}{Type}      & \multicolumn{5}{c}{Setting}                                                                                                                                                                                                                                                                                                                                                                                                & \multirow{2}{*}{Result}   \\ \cmidrule(lr){3-7}
                                                            &                            & Victim Model                                            & Task                                      & Training Dataset                                                                                                                              & Size of Modified Training Data                 & Testing Dataset                                                                                                     &                           \\ \midrule
Carlini et al. \cite{carlini2021poisoning}                  & Poisoning                  & CLIP                                                    & Zero-shot classification                  & \begin{tabular}[c]{@{}c@{}}Conceptual Captions \cite{sharma2018conceptual}\\  or YFCC \cite{thomee2016yfcc100m}\end{tabular}                  & 0.0001\% of training samples                   & ImageNet \cite{deng2009imagenet}                                                                                    & 40\% ASR                  \\
\multirow{2}{*}{Carlini et al. \cite{carlini2023poisoning}} & \multirow{2}{*}{Poisoning} & \multirow{2}{*}{OpenCLIP \cite{ilharco2021openclip}}    & \multirow{2}{*}{Zero-shot classification} & \multirow{2}{*}{LAION-400M \cite{schuhmann2021laion}}                                                                                         & \multirow{2}{*}{0.00025\% of training samples} & \multirow{2}{*}{ImageNet \cite{deng2009imagenet}}                                                                   & \multirow{2}{*}{60\% ASR} \\
                                                            &                            &                                                         &                                           &                                                                                                                                               &                                                &                                                                                                                     &                           \\ 
Nightshade \cite{shan2023prompt}                            & Poisoning                  & Stable Diffusion                                        & Text-to-image generation                  & LAION-5B \cite{schuhmann2022laion}                                                                                                            & 50 samples                                     & \begin{tabular}[c]{@{}c@{}}MSCOCO \cite{lin2014microsoft} and \\ Wikiart dataset \cite{saleh2015large}\end{tabular} & over 70\%  ASR            \\ \midrule
\multirow{2}{*}{Carlini et al. \cite{carlini2021poisoning}} & \multirow{2}{*}{Backdoor}  & \multirow{2}{*}{CLIP}                                   & \multirow{2}{*}{Zero-shot classification} & \multirow{2}{*}{\begin{tabular}[c]{@{}c@{}}Conceptual Captions \cite{sharma2018conceptual} \\ or YFCC \cite{thomee2016yfcc100m}\end{tabular}} & \multirow{2}{*}{0.01\% of training samples}    & \multirow{2}{*}{ImageNet \cite{deng2009imagenet}}                                                                   & \multirow{2}{*}{50\% ASR} \\
                                                            &                            &                                                         &                                           &                                                                                                                                               &                                                &                                                                                                                     &                           \\ 
BadEncoder \cite{9833644}                                   & Backdoor                   & CLIP                                                    & Zero-shot classification                  & CIFAR-10 \cite{krizhevsky2009learning}                                                                                                         & 1\% of training samples                        & SVHN \cite{netzer2011reading}                                                                                       & 100\% ASR                 \\ 
Struppek et al. \cite{struppek2023rickrolling}              & Backdoor                   & Stable Diffusion                                        & Text-to-image generation                  & LAION-Aesthetics \cite{schuhmann2022laion}                                                                                                    & 64 samples per step                            & MSCOCO \cite{lin2014microsoft}                                                                                      & 80 FID                    \\ \midrule
Lapid et al. \cite{lapid2023i}                              & Adversarial                & ViT-GPT2-image-captioning \cite{vitgpt2imagecaptioning} & Image captioning                  & -                                                                                                                                             & -                                              & Flickr30k \cite{plummer2015flickr30k}                                                                               & 0.134 CLIP score          \\ 
RIATIG \cite{Liu_2023_CVPR}                                 & Adversarial                & DALL·E 2                                                & Text-to-image generation                  & -                                                                                                                                             & -                                              & MSCOCO \cite{lin2014microsoft}                                                                                      & 100\% ASR                 \\
Zhuang et al. \cite{Zhuang_2023_CVPR}                       & Adversarial                & Stable Diffusion                                        & Text-to-image generation                  & -                                                                                                                                             & -                                              & Generated by ChatGPT \cite{IntroducingChatGPT}                                                                      & 0.186 CLIP score          \\ \bottomrule
\end{tabular}%
}
\end{table*}

\subsection{Poisoning Attacks} \label{Sec:4.1}\textbf{Challenge.} Among above issues, poisoning attack is one of the most significant and rising concerns for LMs trained by enormous amounts of data acquired from diverse sources, which attempts to lead the model to generate wrong or inappropriate outputs by injecting some poisoning data into the training dataset \cite{huang2023survey}. 

Traditional poisoning attacks are usually performed on classification tasks, and one of the most common ways is to alter the label of training samples \cite{liu_trustworthy_2023,10.1145/1128817.1128824,10.1145/2046684.2046692}. Recently, the work \cite{carlini2023poisoning} showed that it is possible for attackers to poison web-scale datasets like Wikipedia by purchasing website domains or crowdsourcing \cite{liu_trustworthy_2023}. Although it is very difficult to pollute the most part of training data, researchers have shown that poisoning only 0.0001\% of the unlabeled data in self-supervised learning can cause CLIP to significantly misclassify test images \cite{liu_trustworthy_2023,274598,carlini2021poisoning}. When using DMs for image generation, Nightshade \cite{shan2023prompt} fed Stable Diffusion just 50 poisoning images of dogs and then prompted it to create images of dogs itself, the outputs looked like weird creatures with too many limbs and cartoonish faces \cite{Melissa2023poisoning}. In short, the cutting-edge poisoning attack methods can threat the performance of the SOTA models at a small cost. To better evaluate capabilities of poisoning attack methods, a new benchmark called APBench has been proposed to serve as a catalyst for facilitating and promoting future advancements in availability of both poisoning attack and defense methods \cite{qin2023apbench}.

\textbf{Countermeasure.} The core intuition of countermeasures on poisoning attacks is that poisoning data contains malicious information, which leads to abnormality on models, no matter how cleverly to hide them \cite{10.1145/3551636}. Therefore, the most intuitive idea of countering poisoning attacks is to find a metric that can measure the toxicity of data \cite{10.1145/3551636}. For example, Peri et al. \cite{10.1007/978-3-030-66415-2_4} observed that poisoning data have different feature distributions compared with clean samples at higher layers of the neural networks and they proposed a countermeasure to detect poisoning samples in the feature space by Deep K-NN \cite{10.1145/3551636,9431105}. Moreover, many researchers have made efforts on addressing poisoning attacks of label manipulation in different aspects \cite{10.1145/3551636}, e.g. dataset \cite{Li_2017_ICCV}, model architecture \cite{goldberger2017training,Xiao_2015_CVPR}, loss function \cite{Ghosh_Kumar_Sastry_2017,NEURIPS2019_8a1ee9f2,pmlr-v119-liu20e} and optimizer \cite{NIPS2017_58d4d1e7,Patrini_2017_CVPR,NEURIPS2018_ad554d8c}. For pretrained vision-language LMs, it is more economical to use diffusion theory \cite{mandt2017stochastic} to detect model weights related to generating toxic contents and fine-tune it on a small trusted dataset, avoiding training the model from scratch \cite{zhang2023diffusion}. However, the above methods are less effective when encountering more complex poisoning attacks and can only defend against known attack methods. Accordingly, De-Pois \cite{9431105} trains a student model, the purpose of which is to imitate the behavior of the teacher model trained by clean samples, achieving attack-agnostic defense to some extent.

\textbf{Discussion.} Valuable experience can be drawn from the above methods, but it is still an open question to detect or prevent poisoning attacks on prevailing DMs and LMs effectively \cite{10.1145/3551636}.

\subsection{Backdoor Attacks} \label{Sec:4.2}
\textbf{Challenge.} Typical poisoning and backdoor attacks both contaminate the training dataset to mislead model behavior \cite{10.1145/3555803}, while the occurrence of backdoor attacks requires a special signal to trigger. Namely, the goal of backdoor attacks is to cause the model to output a specific pattern when it encounters a specific input, which should not compromise the model performance on clean samples and bypass the human inspection \cite{huang2023survey}. Therefore, backdoor attacks are more stealthy and hard to preclude, which is also reflected in the high similarity between backdoor and benign data \cite{10.1145/3551636}.

Historically, backdoor attacks are introduced on discriminative tasks of light-weight models \cite{8685687,chen2017targeted,10.1145/3372297.3423362,Saha_Subramanya_Pirsiavash_2020}, where the attacker can use a patch/watermark as a trigger and train a backdoored model from scratch \cite{huang2023survey}. Nevertheless, the focus of backdoor attacks has gradually shifted to flourishing LMs. Surprisingly, by maliciously modifying just 0.01\% of training data, it is possible to make CLIP misclassify test images to any desired label by masking a small patch of the test images \cite{carlini2021poisoning}. Specifically, BadEncoder\cite{9833644} was proposed to fine-tune a backdoored image encoder from a clean one and achieves high attack success rates while preserving the accuracy of the downstream classifiers.

Likewise, the investigation into backdoor attacks has extended to DMs \cite{Chen_2023_CVPR,Chou_2023_CVPR,struppek2023rickrolling,huang2023zeroday,zhai2023texttoimage,wang2023robust,vice2023bagm}. For instance, Struppek et al. \cite{struppek2023rickrolling} employs a teacher-student approach to integrate the backdoor into the pretrained text encoder and validate the backdoor attacks, e.g. Latin characters are replaced with similar-appearing Cyrillic trigger characters, and the generated images will follow a specific description or include certain attributes \cite{huang2023survey}. To better measure backdoor attack methods, a unified evaluating framework for backdoor attacks in DMs has been proposed, which facilitates the backdoor analysis of different task settings and provides new insights into caption-based backdoor attacks on DMs \cite{chou2023villandiffusion}.

\textbf{Countermeasure.} The main goal of countermeasures against backdoor attacks is detection, which can be carried on the whole life-cycle of a model, including data collection, training stage, and inference stage \cite{10.1145/3551636}. For example, PatchSearch \cite{Tejankar_2023_CVPR} proposes to train a model from scratch based on a secure dataset, which is filtered by a pretrained model. Likewise, Bansal et al. \cite{bansal2023cleanclip} hold the view that learning representations for each modality independently from the other could break correlation between the backdoor trigger and the target label, and propose CleanCLIP framework to fine-tune the pretrained CLIP on a clean dataset. Furthermore, SAFECLIP \cite{yang2023better} has been proposed to warms up the model by applying unimodal contrastive learning on image and text modalities separately, which then carefully divides the multimodal data into safe and risky subsets to further train the model. Furthermore, DECREE \cite{Feng_2023_CVPR} first searches for a minimal trigger pattern such that any inputs stamped with the trigger share similar embeddings, which is then utilized to decide whether the given encoder is benign or trojaned. Intuitively, knowledge distillation can also be used to defend against backdoor attacks by distilling away the trigger \cite{Saha_2022_CVPR}.

\begin{figure}[ht]
    \centering
    \includegraphics[width=\linewidth,page=8]{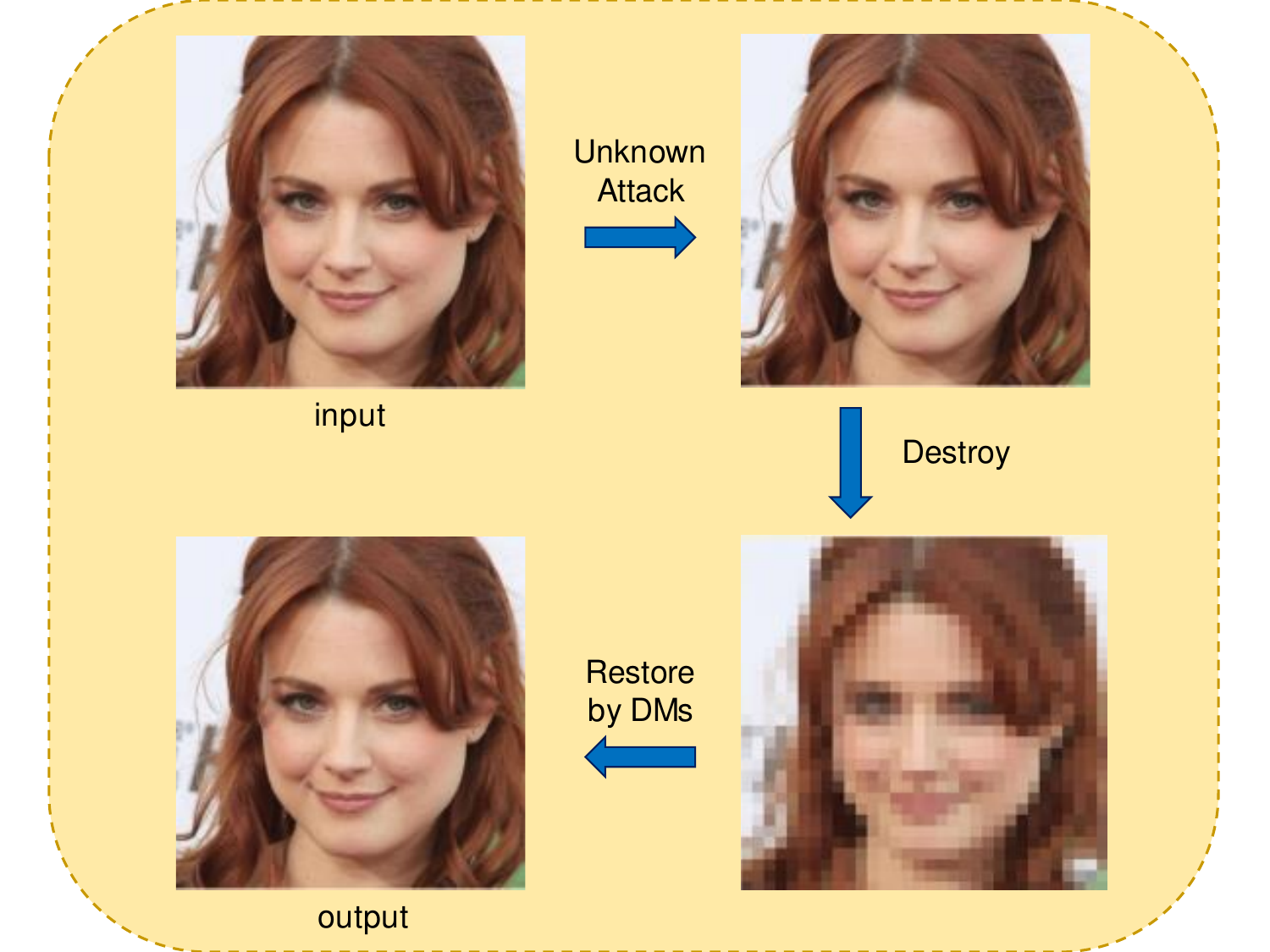}
    \caption{The summary of using DMs to deal with backdoor attacks by restoring the poisoning image. The figure is adapted from the work \cite{shi2023blackbox}.}
    \label{BA}
\end{figure}

For generative tasks, one way to defend backdoor attack is to create synthetic variations of all training samples, leveraging the inherent resilience of DMs to trigger potential patterns in the data \cite{struppek2023leveraging}. By further combining this method with knowledge distillation, student models can maintain their general performance on the task, and meanwhile exhibit robust resistance to backdoor attacks \cite{struppek2023leveraging}. Interestingly, DMs themselves can be used to defend against backdoor attacks by reconstructing the image \cite{may2023salient,shi2023blackbox}, as shown as Figure \ref{BA}. 

\textbf{Discussion.} Backdoor attacks pose a critical threat to model performance, which becomes even more severe since the models' memory capacity continue to improve \cite{fan2023trustworthiness}. Specifically, the expansive capacity of these models produces ample space to establish an association between triggers and target behaviors, even with little poisoning data \cite{fan2023trustworthiness}. Backdoor attacks could cause catastrophic damage to downstream applications that depend on the LMs, while research on methods to effectively combat backdoor attacks is still limited \cite{chen2023pathway}. 


\subsection{Adversarial Attacks} \label{Sec:4.3}
\begin{figure}[ht]
  \centering
  \includegraphics[width=\linewidth,page=6]{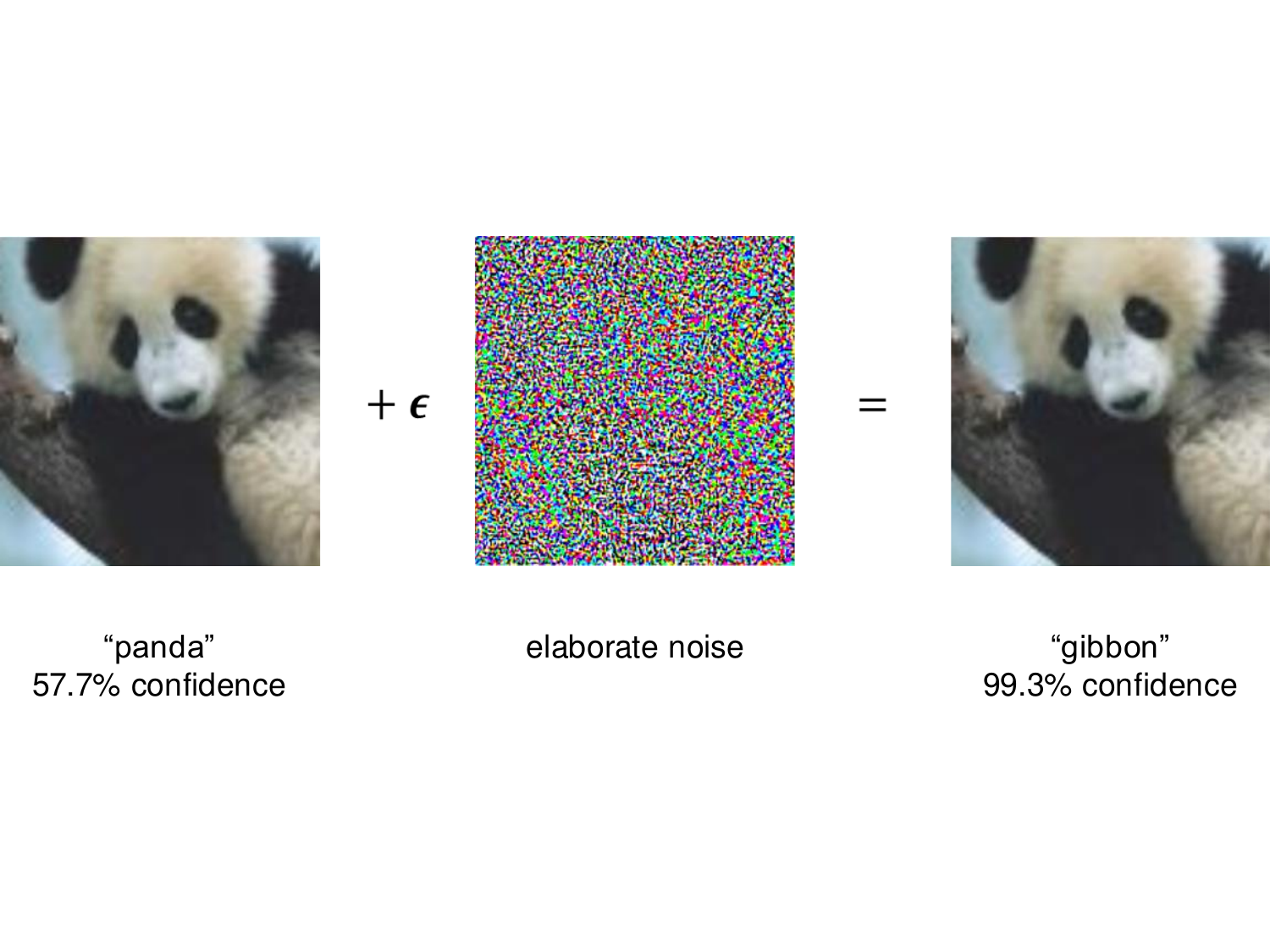}
  \caption{One of the most well-known example of adversarial attacks in computer vision, which is adapted from the work \cite{goodfellow2014explaining}. It shows that simple modifications to the input image can mislead the model to make incorrect predictions. }
  \label{AA}
\end{figure}

\textbf{Challenge.} The last widely studied topic under model vulnerability is adversarial attacks, which mainly study the model’s reaction to imperceptible and malicious input when attackers do not have the access to training dataset compared to other attacks. Since most users do not have access to models' training data, adversarial attacks are the easiest attack to implement among three attacks, which directly exploit inherent vulnerabilities in models \cite{fan2023trustworthiness}. The initial focus of adversarial attacks was on misleading models, where an offset is added to the original input as the adversarial examples \cite{goodfellow2014explaining}, as shown as Figure \ref{AA}.  In the context of conventional image classification tasks, a well-known example involves injecting sophisticated noise into an image. This results in an image that outwardly resembles the original but misleads the model into predicting it as something different \cite{liu2023towards}. In addition to decreasing model performance, adversarial attacks can also increase the computational cost of models. During model training, direction and magnitude of the gradients can be maliciously modified, causing larger computational cost during inference \cite{Haque_2020_CVPR,pan2022gradauto,pan2023gradmdm}.

A series of studies on adversarial attacks of LMs have also emerged, mainly including visual adversarial attacks \cite{lapid2023i} and text adversarial attacks \cite{Liu_2023_CVPR,millière2022adversarial}. In image caption, visual adversarial attacks may be able to cheat CLIP and BLIP to generate conversations unrelated to input images by injecting learnable noise into images \cite{zhao2023evaluating}. Surprisingly, these attacks can be transferred to other models like MiniGPT-4 \cite{zhu2023minigpt4} or Large Language and Vision Assistant (LLaVA) \cite{touvron2023llama,liu2023visual} and can already induce targeted responses with a high success rate \cite{zhao2023evaluating}. In text-to-image generation, untargeted textual attack and targeted textual attack have been introduced to generate unrelated or prescribed images respectively, which roots in the lack of robustness of text encoders \cite{Zhuang_2023_CVPR}.  Moreover, automatic dataset construction pipelines for textual adversarial attacks have been proposed \cite{zhang2023robustness,maus2023black}.

\textbf{Countermeasure.} While long lines of studies on adversarial attacks have been conducted \cite{liu2023towards,10.1007/978-3-030-58592-1_29,Dong_2018_CVPR,pmlr-v97-guo19a}, Projected Gradient Descent (PGD) \cite{madry2017towards} is considered as the most powerful defence methods. Intuitively speaking, this method is to perform gradient ascent on adversarial examples before gradient descent for learning representation, because it can find the most effective adversarial examples for learning \cite{liu_trustworthy_2023,gan2020large}. However, PGD ignores the visual-language alignment in CLIP's pretrained representation space, causing the model to lose zero-shot capability \cite{mao2022understanding}. Thus, a text-guided contrastive adversarial training loss, TeCoA, has been proposed, which maximizes the similarity of the adversarial visual features and the corresponding text embeddings in contrastive learning \cite{mao2022understanding}. 

\textbf{Discussion.} Overall, in the field of adversarial attacks in LMs, many of the aforementioned defence methods rely on white-box models while attack methods can be implemented on black-box models. Thus, the development of defence methods still lags far behind that of attacking methods, which hinders the widespread application of LMs. This endless arms race between attacks and defences, emphasizes the necessity of incorporating security measures of adversarial attacks from the outset \cite{fan2023trustworthiness}.

\section{Inherent Issue} \label{Sec:5}
\begin{figure}[ht]
  \centering
  \includegraphics[width=\linewidth,page=3]{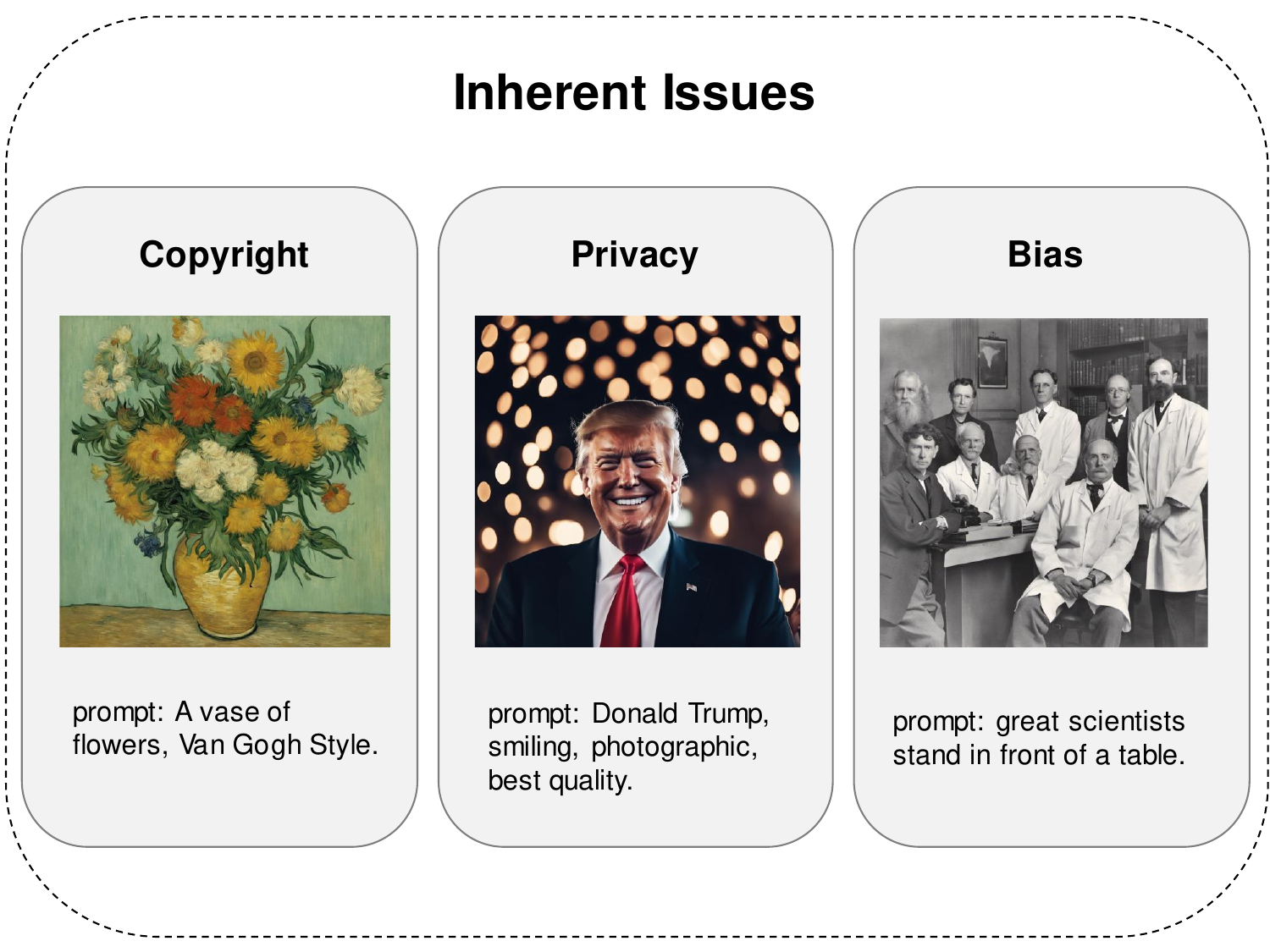}
  \caption{Examples of the inherent issue, showing copyright plagiarism in the style of the artist Van Gogh, privacy violations related to political celebrity Trump, and gender bias towards females, which are generated by Stable Diffusion XL v1.0 in November 2023. }
  \label{II}
\end{figure}

Although the behavior of LMs has been imposed some constraints, they still face many deep-rooted problems to accommodate to human trust, namely inherent issues. Distinct from human misuse caused by human subjectivity, inherent issues refers to the intrinsic issues of LMs, which are not related to the malicious behavior of the user. In contrast, inherent issues generally occur beyond the user's expectations. We will further discuss major challenges in inherent issues below including Copyright in Section \ref{Sec:5.1}, Privacy in Section \ref{Sec:5.2}, Bias in Section \ref{Sec:5.3} and Hallucination in Section \ref{Sec:5.4}. Some corresponding examples are shown in Figure \ref{II}, Table \ref{IIT} and Figure \ref{OB}.

\begin{table*}[]
\caption{A summary of inherent issues of trustworthy LLMs in vision.}
\label{IIT}
\resizebox{\textwidth}{!}{%
\begin{tabular}{@{}cccccc@{}}
\toprule
\multirow{2}{*}{Model} & \multirow{2}{*}{Task}                                                   & \multicolumn{4}{c}{Challenge}                                                                                                                                                                                                                                                                                                                                                                                                                                           \\ \cmidrule(l){3-6} 
                       &                                                                           & Copyright                                        & Privacy                                                                                   & Bias                                                                                                                                                       & Hallucination                                                                                                                                               \\ \midrule
GPT-4                   & \begin{tabular}[c]{@{}c@{}}Image-to-text \\ and text-to-text\end{tabular} & -                                                & -                                                                                         & -                                                                                                                                                          & \begin{tabular}[c]{@{}c@{}}Generate captions containing\\  nonexistent or inaccurate \\ objects from the input image \cite{liu2023hallusionbench}\end{tabular} \\ 
DALL·E series          & Text-to-image                                                             & Replicate training images \cite{OpeanAIDALLE2}   & -                                                                                         & \begin{tabular}[c]{@{}c@{}}Reinforce racial stereotypes like \\ combining negative adjectives with \\ non-white people \cite{Khari2023DALLE2}\end{tabular} & -                                                                                                                                                           \\
Stable Diffusion       & Text-to-image                                                             & Imitate human artists \cite{casper2023measuring} & \begin{tabular}[c]{@{}c@{}}Output trademarked\\  company logos \cite{291199}\end{tabular} & \begin{tabular}[c]{@{}c@{}}Generate images of white men with positive \\ words by default \cite{chen2023pathway}\end{tabular}                              & -                                                                                                                                                           \\ \bottomrule
\end{tabular}%
}
\end{table*}

\subsection{Copyright}\label{Sec:5.1}
\textbf{Challenge.} In order to generate high-quality content, LMs require substantial quantities of training data, which may often involve information protected by copyright and poses a threat to intellectual property rights \cite{lucchi_2023}. For instance, textual inversion \cite{gal2022image}, a novel method based on the latent diffusion model, can imitate the art style of human-created paintings with several samples, which extremely infringes on the intellectual property of artists \cite{liang2023adversarial}. Furthermore, LMs like DALL·E 2 \cite{rameshHierarchicalTextConditionalImage2022} is trained on billions of image-text pairs from the Internet, the size of which is too large to check whether each containing pair is authorized or not. OpenAI researchers also found that DALL·E 2 occasionally replicated training images, as the training dataset contained a substantial number of duplicated images \cite{OpeanAIDALLE2}. What's even worse is that, due to the memorization ability of LMs \cite{291199,carlini2022quantifying,tirumala2022memorization,zhang2021counterfactual,jagielski2022measuring}, users are able to directly extract copyright-protected content from LMs and cause copyright infringement \cite{OpeanAIDALLE2}.

\textbf{Countermeasure.}  Many works have emerged to alleviate the problem of copyright infringement in LMs. For instance, OpenAI researchers use K-Means clustering \cite{8ddb7f85-9a8c-3829-b04e-0476a67eb0fd} to remove simple and duplicate images before pretraining and achieved better results in human evaluation \cite{OpeanAIDALLE2}. From the perspective of training, models can also be encouraged to generate novel images, which are far away from original images in the training dataset \cite{vyas2023provable}, while deleting repeated images reduces the size of training dataset. In addition, adversarial attack is also a notable method for protecting the copyright of images. Specifically, perturbations are added to target images to safeguard data against unauthorized training, disrupting the model's ability to learn the corresponding representations, and has achieved significant results \cite{liang2023adversarial,shan2023glaze,wang2023plugandplay,cui2023diffusionshield, 10097580, guo2023domain,Van_Le_2023_ICCV,salman2023raising}. Interestingly, for the sake of protecting copyrighted artistic styles, model fine-tuning enables DMs to erase specific artistic styles in generated outputs \cite{gandikota2023erasing,gandikota2023unified,kumari2023ablating}. 

\textbf{Discussion.} Copyright protection has gradually become one of the primary concerns in the field of LMs, as there is no strong performance of LMs without a large amount of available and authorized data. More research is needed to safeguard copyright and mitigate the risk of copyright infringement in the domain of LMs. 

\begin{table*}[ht] 
\caption{\label{Bias} 
Comparison of five major countermeasures to mitigate privacy leakage.
}

\begin{center}
\resizebox{\linewidth}{!}{
\begin{tabular}{ccc}

\toprule 
Countermeasure & Advantage & Limitation \\ 
\midrule 

Deleting duplicate data \cite{OpeanAIDALLE2} & Straightforward and simple & Only applicable to scenarios with duplicate data \\ 

Data anonymization \cite{maximov2020ciagan} & Applicable to most scenarios & High computational cost \\ 

Adversarial attack \cite{hintersdorf2023defending} & Simple and applicable to most scenarios & Could be invalidated by defence methods \\ 

Differential privacy \cite{dwork2006differential,abadi2016deep} & Applicable to most scenarios & Increase the training cost and hurt utility of models \\ 

Machine unlearning \cite{ginart2019making, nguyen2020variational, canteaut2020advances, xu2023machine, krishnan2017palm} & Avoid training from scratch & Hard to verify validity  \\ 

\bottomrule 
\end{tabular}}
\end{center}
\end{table*}

\subsection{Privacy}\label{Sec:5.2}
\textbf{Challenge.}  The lately released AI ethics guidelines \cite{10.1145/3555803, cannarsa2021ethics, jobin2019global, hagendorff2020ethics, siau2020artificial, whittlestone2019role, WHITEFACT2023} increasingly highlight privacy concerns because sensitive user data are always collected to further train LMs causing privacy leakages \cite{foo_aigc_2023}. Such privacy leakages in DMs are recently well-documented \cite{291199,Somepalli_2023_CVPR,dar2023investigating} and raise pursuit of privacy protection \cite{liu_trustworthy_2023}. Furthermore, Membership Inference Attacks \cite{shokri2017membership}, where the attacker can determine whether an input instance is in the training dataset or not, have been conducted on DMs \cite{duan2023diffusion,zhu2023data,huang2023survey}. Due to burdensome computation of directly extracting data from models, it seems more feasible to conduct membership inference attacks on these models \cite{fan2023trustworthiness}. There is also an unexpected usage of membership inference attacks, which is to detect whether model training had illegally utilized the unauthorized data \cite{NEURIPS2022_55bfedfd, fan2023trustworthiness,wang2023detect}.

\textbf{Countermeasure.} To reduce the risk of privacy leakage, commonly used countermeasures can be categorized into five main types \cite{fan2023trustworthiness}, which is shown in Table \ref{Bias}. The first method aims at deleting duplicate data from training dataset as mentioned in Section \ref{Sec:5.1}. 
The alternative method, data anonymization, subtly modifies the data so that the protected private information cannot be recovered \cite{10.1145/3555803}. Data anonymization has been applied to reduce the risk of privacy leakage \cite{10.1145/3555803,maximov2020ciagan,peng2023joint,Li_2023_ICCV,10230346,Klemp_2023_CVPR}, which is anticipated to be resistant to the recovery of private information \cite{el2011systematic, ji2014structural, 10.1145/3555803}. From an alternative perspective, vulnerability of LMs can also be used to protect privacy. More specifically, by injecting noises into training data, e.g. performing adversarial attacks, models are prevented from learning effective representation from training data \cite{hintersdorf2023defending}. 
Another method, differential privacy \cite{dwork2006differential,abadi2016deep}, intends to prevent the model from over-memorizing specific information associated with individual data points by injecting noises into gradients during training \cite{song2013stochastic, fan2023trustworthiness}. However, differential privacy could significantly increase the training cost and hurt the utility of models, because directions of stochastic gradient descent are not optimal due to injection of noises \cite{carlini2019secret,carlini2021extracting, deng2023recent}.

The last method, machine unlearning, aims to achieve that the influence of target training samples can be completely and quickly removed from the trained model, which is motivated by that removing the influence of outlier training samples from a model will lead to higher model performance and robustness \cite{ginart2019making, nguyen2020variational, canteaut2020advances, xu2023machine, krishnan2017palm}. According to the design details of the unlearning strategy, it can be divided into two categories, namely Data Reorganization and Model Manipulation \cite{xu2023machine}. Although machine unlearning can remove the knowledge about target data in the trained model without training the model from scratch in theory, how to verify whether the model has ``forgotten'' the target data is still a problem, especially when the size of training dataset is particularly huge \cite{xu2023machine}. To mitigate this issue, empirical evaluation and theoretical calculation have been proposed to verify effectiveness of unlearning \cite{xu2023machine}.

\textbf{Discussion.} In conclusion, existing measures to protect privacy are inadequate to meet the demands \cite{chen2023pathway}. It can even be said that investigation of privacy defenses is noticeably lagging behind that of the attacks and further investigation is needed for the optimal trade-off between model utility and privacy protection \cite{fan2023trustworthiness}.

\subsection{Bias}\label{Sec:5.3}
\textbf{Challenge.} Bias is associated with an unjustified negative stance toward a social group, stemming from one-sided or inaccurate information, which typically involves widely circulated negative stereotypes pertaining to factors such as gender, race, religion, and more \cite{deng2023recent, sekaquaptewa2003stereotypic}. Due to the nature of training on uncurated training data, LMs have gained negative reputations for generating negative contents instilling and exacerbating social bias \cite{liu_trustworthy_2023, deng2023recent, 10.1145/3555803}. For example, a recent work \cite{Khari2023DALLE2} shows that DALL·E 2 leans toward generating images of white men with positive words by default, overly sexualizes images of women like binding females to flight attendants, and reinforces racial stereotypes like combining negative adjectives with non-white people. Even a simple synonym replacement will cause DALL·E 2 to induce cultural stereotypes in the generated images \cite{struppek2022exploiting}. However, to guarantee that the generated content is unbiased to different groups is particularly important, contributing to measuring the social impact of LMs and promoting its safer deployment \cite{deng2023recent}.

\begin{figure}[ht]
  \centering
  \includegraphics[width=\linewidth,page=11]{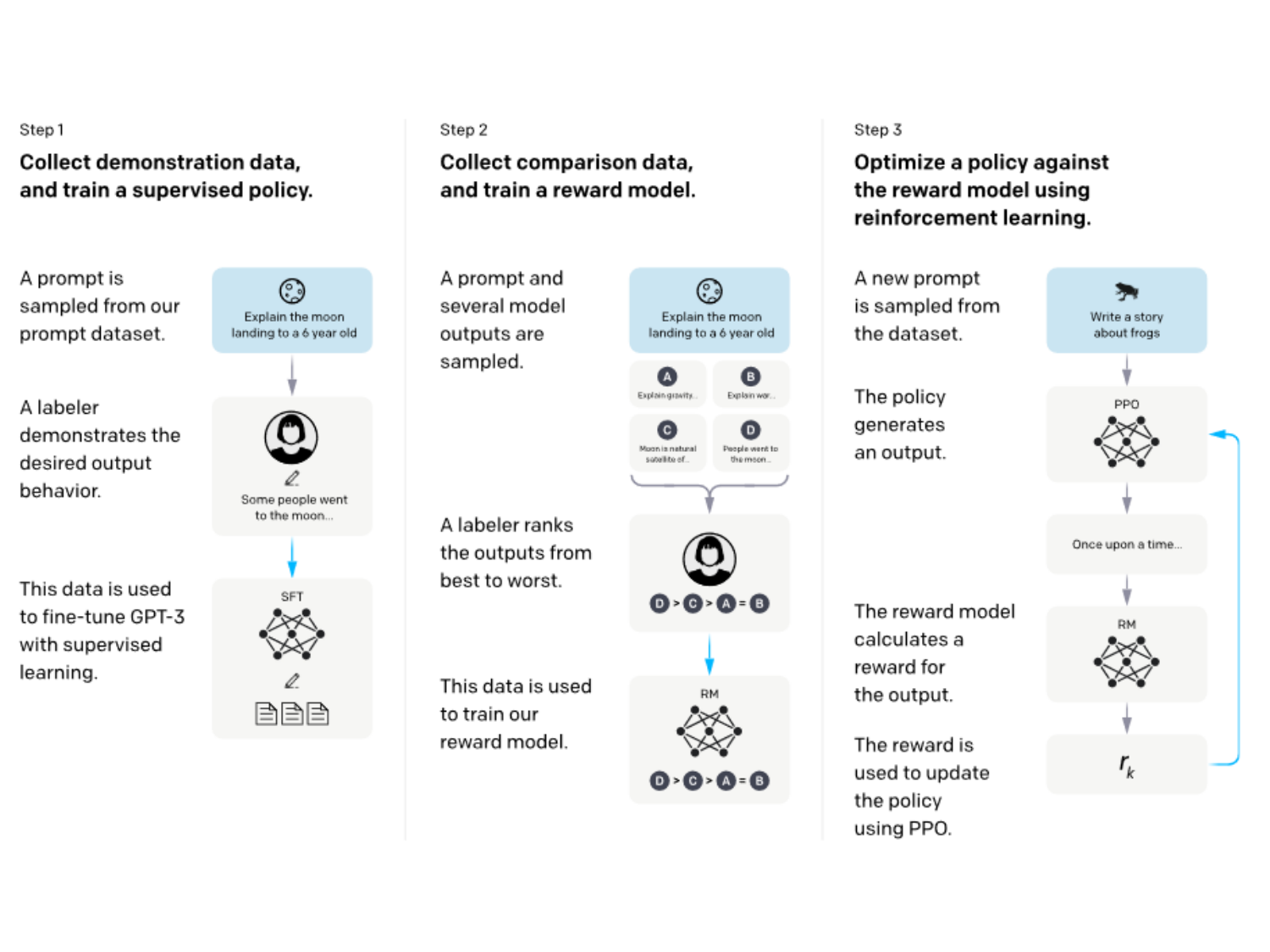}
  \caption{The general process of RLHF, including (1) supervised fine-tuning (SFT), (2) reward model (RM) training, and (3) reinforcement learning via proximal policy optimization (PPO). The figure is from the work \cite{ouyang2022training}.}
  \label{RLHF}
\end{figure}

\textbf{Countermeasure.} As the quality of the content generated by LMs is inextricably linked to the quality of the training dataset, the most straightforward and effective approach to address this issue is to eliminate toxic data from the training dataset \cite{fan2023trustworthiness, chen2023pathway}. For instance, OpenAI researchers filter toxic and biased content from training dataset \cite{OpeanAIDALLE2}. Furthermore, they further re-weight the imbalanced data in loss function to rectify bias of models and significantly improve the model performance on a metric for evaluating the balance of different labels \cite{OpeanAIDALLE2}. 

On the other hand, Reinforcement Learning from Human Feedback (RLHF) \cite{ouyang2022training} has been widely applied in training models for text generation aligned with human preference and achieved great success, e.g. it has been used in fine-tuning GPT-3 and GPT-4 for various scenarios \cite{GPT-4-Technical-Report}, as shown in Figure \ref{RLHF}. In a similar vein, RLHF can also be applied to training models for image generation. For example, the model architecture and training pipeline of the DMs can be modified to align with human expectations by RLHF \cite{lee2023aligning,xu2023imagereward}. Accordingly, RLHF has the potential to guide the LMs to generate content with less bias \cite{bai2022constitutional}. Intuitively, the correction mechanism for bias can be directly deployed in the inference stage. The methods for addressing model bias in text generation \cite{dathathri2019plug,krause2020gedi,laugier2021civil,schick2021self} can also be adapted to detect and filter out biased output in visual tasks in theory \cite{gandikota2023unified}. 

\textbf{Discussion.} In general, the flourish of LMs drives bias to be a concern of importance and efforts should be made to overcome bias for trustworthy LMs. Although measures such as data filtering and RLHF can help alleviate this concern, they are not foolproof \cite{fan2023trustworthiness}. Therefore, there is still a long way to go in evaluating and rectifying bias during the life-cycle of LMs' training and development \cite{chen2023pathway}.

\subsection{Hallucination} \label{Sec:5.4}
\begin{figure}[ht]
  \centering
  \includegraphics[width=\linewidth,page=4]{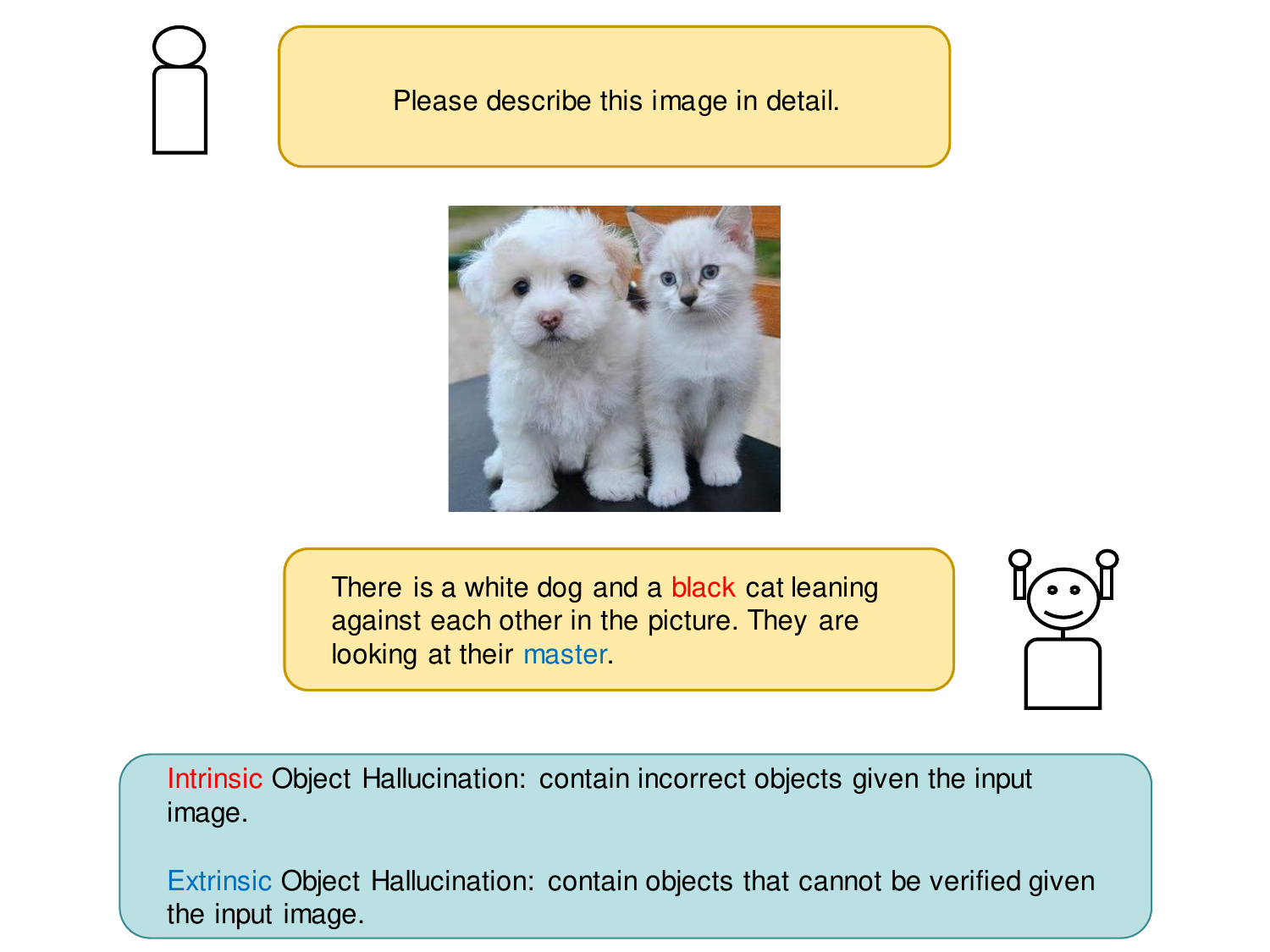}
  \caption{Examples of intrinsic and extrinsic object hallucination in LMs. The image of animals is from the work \cite{NEURIPS2022_a1859deb}.}
  \label{OB}
\end{figure}

\textbf{Challenge.} In the domain of Natural Language Processing (NLP), the term ``hallucination'' describes a phenomenon where the content generated by models lacks coherence or deviates significantly from the provided source content, despite being expressed with apparent confidence \cite{10.1145/3571730,filippova2020controlled,maynez2020faithfulness,parikh2020totto,zhao2021ror}. However, as the emergence of vision-language LMs like CLIP and BLIP, hallucination has also been observed in field of CV and poses a great risk to performance of LMs. In the domain of vision-language LMs, this significant challenge is further defined as object hallucination \cite{rohrbach2018object}, where generated captions contain nonexistent or inaccurate objects from the input image \cite{10.1145/3571730}. To be more specific, object hallucination can be more precisely categorized into intrinsic object hallucination and extrinsic object hallucination \cite{10.1145/3571730}, the details of which are shown in Figure \ref{OB}. Moreover, hallucinations are not always harmful and can provide positive pairs for self-supervised contrastive learning, which are able to improve models’ accuracy \cite{Wu_2023_ICCV}.

\begin{table*}[ht] 
\caption{\label{hallucination_metric}Comparison of several types of metrics for evaluating detection of object hallucination.}
\begin{center}

\resizebox{\linewidth}{!}{

\begin{tabular}{ccc}
\toprule 
Metric & Advantage & Limitation \\ 
\midrule 

BLEU \cite{papineni2002bleu} & Automatically calculated & Correlate weakly with human judgment \\ 

ROUGE \cite{lin2004rouge} & Automatically calculated & Correlate weakly with human judgment \\ 

METEOR \cite{banerjee-lavie-2005-meteor} & Incorporate recall factor and explicit word-matching between translation and reference & Not faithful to reflect hallucination \\ 

CIDEr \cite{Vedantam_2015_CVPR} & More straightforward and objective  & Not faithful to reflect hallucination \\ 

SPICE \cite{10.1007/978-3-319-46454-1_24} & Overcome limitations of n-gram  & Not faithful to reflect hallucination \\ 

CHAIR \cite{rohrbach2018object} & More relevant to human ratings  & Unrobust and need complex human-crafted parsing rules \\ 

\bottomrule 
\end{tabular}}
\end{center}
\end{table*}

\begin{table*}[ht] 
\caption{\label{hallucination_benchmark} Comparison of several types of benchmarks for evaluating detection of object hallucination.}
\begin{center}

\resizebox{\linewidth}{!}{

\begin{tabular}{ccc}
\toprule 
Evaluation & Advantage & Limitation\\ 
\midrule 

POPE \cite{li2023evaluating} & More stable and flexible & Not suitable for complex real-world scenarios \\ 

HaELM \cite{wang2023evaluation} & Low cost, reproducible, privacy preservation & LLMs-based approach may itself creates hallucination \\ 

GAVIE \cite{liu2023mitigating} & Independent on groundtruth answers & LMs-based approach may themselves create hallucination \\ 

MMHAL-BENCH \cite{sun2023aligning} & More general, realistic, and open-end & LMs-based approach may itself create hallucination \\ 

HALLUSIONBENCH \cite{liu2023hallusionbench} & Having more abundant and fine-grained topics & LMs-based approach may itself create hallucination \\ 

\bottomrule 
\end{tabular}}
\end{center}
\end{table*}

\textbf{Evaluation.} To measure the extent of object hallucination, it was a common practice to rely on automatic sentence metrics, e.g. BLEU \cite{papineni2002bleu}, ROUGE \cite{lin2004rouge}, METEOR \cite{banerjee-lavie-2005-meteor}, CIDEr \cite{Vedantam_2015_CVPR} and SPICE \cite{10.1007/978-3-319-46454-1_24} to evaluate captioning performance during deployment \cite{rohrbach2018object}, most of which has been shown in Table \ref{hallucination_metric} and Table \ref{hallucination_benchmark}. Yet the above metrics are less relevant to human ratings or cannot reflect the existence of hallucinations faithfully because captions with hallucination can still have good scores as long as they contain sufficient accurate objects to fulfill coverage \cite{rohrbach2018object,dai2022plausible}. Consequently, 
Caption Hallucination Assessment with Image Relevance (CHAIR) \cite{rohrbach2018object} metric has been proposed to be more relevant to human ratings, which calculates what proportion of generated object words are actually in the image according to the ground truth captions \cite{10.1145/3571730}. Nevertheless, CHAIR can be affected by factors such as instruction designs and the length of captions, which indicates the instability of the CHAIR metric \cite{li2023evaluating}. In addition to instability, CHAIR metric needs complex human-crafted parsing rules to perform exact matching, which have not been adapted to the special generation styles of LMs \cite{li2023evaluating}. In order to stably and conveniently evaluate the object hallucination, Polling-based Object Probing Evaluation (POPE) \cite{li2023evaluating} has been proposed, which firstly employs an object detector to identify all objects within an image and subsequently utilizes predefined prompts to query the model if an object exist in the image. However, POPE is only suitable for ideal situations and is not suitable for complex real-world scenarios, indicating that POPE cannot be used as an excellent basis for hallucination evaluation \cite{wang2023evaluation}. Therefore, an innovative framework called Hallucination Evaluation based on Large Language Models (HaELM) \cite{wang2023evaluation} has been proposed, which evaluates object hallucination by fine-tuning LMs through LoRA \cite{hu2021lora}. While HaELM has additional advantages including low cost, reproducibility, privacy preservation and local deployment, it still falls short of achieving human-level hallucination evaluation performance \cite{wang2023evaluation}. By fully leveraging the power of LMs, GPT4-Assisted Visual Instruction Evaluation (GAVIE) \cite{liu2023mitigating} has been proposed to evaluate the object hallucination in a training-free manner, which serves as an human expert and does not rely on human-annotated groundtruth answers. Moreover, to further quantify and evaluate the object hallucination, a new benchmark MMHAL-BENCH \cite{sun2023aligning} containing 8 types of questions and 12 object categories has been proposed, the evaluation standard of which is more general, realistic, and open-ended compared to previous benchmarks. Similar to MMHAL-BENCH, a new benchmark called HALLUSIONBENCH \cite{liu2023hallusionbench} has been introduced as the first benchmark focusing on the visual and knowledge hallucination of LMs, covering topics ranging from analysing charts to reasoning math. 

\textbf{Countermeasure.} To tackle the issue of object hallucination, a Multimodal Hallucination Detection Dataset (MHalDetect) has been introduced, which can be used to train and benchmark models for detecting and preventing hallucination \cite{gunjal2023detecting}. To demonstrate the potential of MHalDetect for preventing hallucination, a novel Fine-grained Direct Preference Optimization (FDPO) has been applied in optimizing models, which significantly reduces multimodal hallucinations \cite{gunjal2023detecting}. 
In addition, from the data level, three simple yet efficient data augmentation methods have been proposed to mitigate object hallucination, which requires no new training dataset or increase in the model size \cite{Biten_2022_WACV}. 

From the level of training models, Self-Critical Sequence Training (SCST) \cite{Rennie_2017_CVPR} was the most representative measure to significantly boost image caption quality. However, directly fine-tuning models with SCST strategy causes over-fitting on the fine-tuning dataset and leads to object hallucination \cite{dai2022plausible,zhai2023investigating}. Similar to that RLHF can be used to improve model performance in domain of NLP, Factually Augmented RLHF \cite{sun2023aligning} has been proposed to improve model generalization ability and better align model output with human expectations in domain of vision-language. Besides, a new dataset called LRV-Instruction \cite{liu2023mitigating} has been proposed to fine-tune LMs like MiniGPT4 and mPLUG-Owl \cite{ye2023mplug} to reduce hallucination and improve models' performance on several public datasets compared to SOTA methods. On the other hand, some argue that the cause of object hallucination lies in the misalignment between the image and text \cite{dai2022plausible, sun2023aligning}. To mitigate this problem, the method that replaces the previous loss function with ObjMLM loss \cite{dai2022plausible} in pretraining to encourage alignment between different modalities has been proposed. For tasks involving next token predicting, an uncertainty-aware beam search method \cite{xiao2021hallucination} has been proposed for less uncertainty and hallucination in model output, where a weighted penalty term is added to the beam search objective to balance between log probability and predictive uncertainty of the selected token candidates \cite{10.1145/3571730}. 

\textbf{Discussion.} In general, object hallucinations remain deeply rooted in LMs, and their causes are not yet fully explained. Therefore, there is still an urgent need to ensure that LMs produce accurate and trustworthy content \cite{rawte2023survey}.

\section{Interpretability}
\textbf{Challenge.} Another big branch of the study of trustworthy LMs is the investigation to unveil the blackbox nature of LMs, aiming to explain its reasoning mechanisms and provide insights into how it generates content \cite{liu_trustworthy_2023,liu2023towards}. This field is often described as to study the explanability and interpretability of LMs \cite{liu2023towards}, but there is often confusion between the meanings of these two words. The aim of studying explainability is to elucidate how deep neural networks can employ gradient descent for representation learning, while studying interpretability aims to assess how deep neural networks, particularly LMs, can establish causal relationships between input and output during the inference process \cite{Rudin2019Stop}. Nevertheless, current approaches remain unable to comprehensively explain intricate LMs, and researchers have turned to explore post-hoc interpretability methods \cite{10.1145/3555803}, which will be further discussed below.

Interpretability, i.e. understanding how LMs make decision, stays at the core place of cutting-edge LMs research and serves as a fundamental factor that affects the trust in LMs. Despite the prevalence of LMs in various application scenarios, the opacity of complex LMs has raised significant concerns in academia, industry, and society at large \cite{10.1145/3555803}. For instance, lack of interpretability, trust and transparency creates a barrier to the application of LMs in clinical practice, even they rival or exceed the performance of human experts \cite{Teng2022interpretability}. In addition, social and cultural biases are introduced and amplified at many stages of DALL·E 2 development and deployment \cite{Khari2023DALLE2}, while how the biases are propagated remains unclear \cite{chen2023pathway}. As a result, trustworthy service of LMs may face significant obstacles if users do not comprehend how the output is generated by LMs \cite{liu_trustworthy_2023}. 

\begin{figure}[ht]
  \centering
  \includegraphics[width=\linewidth,page=10]{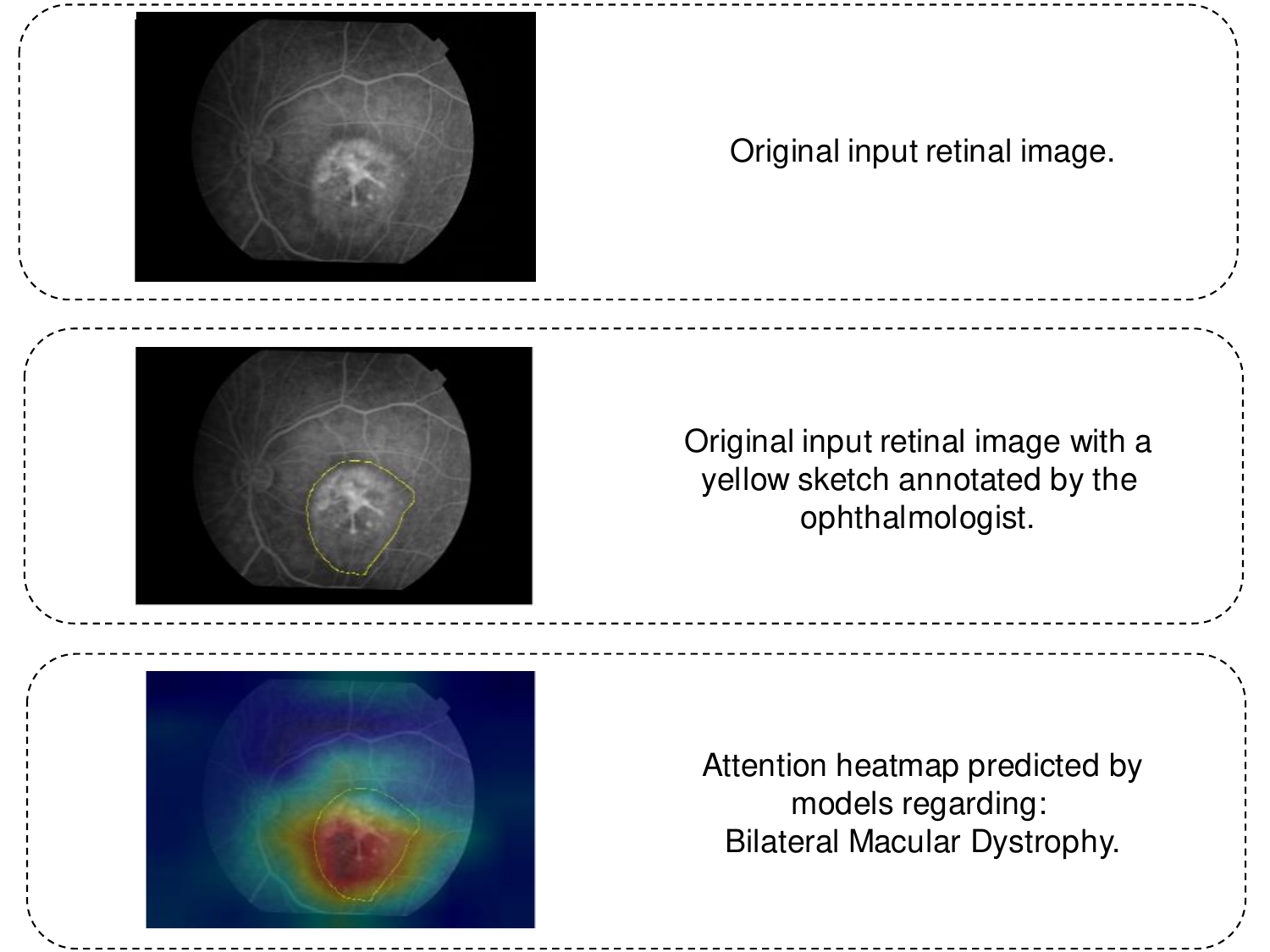}
  \caption{A heatmap based on Grad-CAM \cite{Selvaraju_2017_ICCV}, the red regions of which indicate the most related areas to expert-defined keywords: Bilateral Macular Dystrophy. The figure is adapted from the work \cite{Wu_2023_WACV}.}
  \label{CAM}
\end{figure}

\textbf{Evaluation.} Previous assessments of visual interpretability primarily relied on heat maps showing the difference in importance of each image region, such as Grad-CAM \cite{Selvaraju_2017_ICCV} and BagNet \cite{brendel2019approximating}. In addition, there exist methods based on finding prototypes such as ProtoPNet \cite{NEURIPS2019_adf7ee2d} and ProtoTree \cite{Nauta_2021_CVPR}. Such methods provide interpretability of models by directly comparing the details between training images and test images to get scores for models' judgement.

\textbf{Countermeasure.} Studying interpretability of LMs can draw on previous solutions for visual models and can be further facilitated by leveraging information from language modality. For instance, in the field of image medical diagnosis, expert-defined keywords are effective to indicate diagnosis areas combined with attention mechanism \cite{Wu_2023_WACV}. To be more specfic, this work \cite{Wu_2023_WACV} proposes to exploit such keywords and a specialized attention-based strategy to build a more human-comprehensible medical report generation system for retinal images, as shown in Figure \ref{CAM}. In order to provide visual interpretability of LMs predictions, there exists the work \cite{li2022exploring} proposing the Image-Text Similarity Map to further explain the raw predictions of LMs. In the context of image classification, it is more rational to task LMs with identifying descriptive features rather than direct classification. For instance, to locate a tiger, the model should focus on recognizing its stripes, claws, and other specific attributes, thereby enhancing interpretability \cite{menon2022visual,Yan_2023_ICCV,Yang_2023_CVPR}. In image-text retrieval, image captioning, and visual question answering tasks, a study \cite{Palit_2023_ICCV} suggests the corrosion of certain image tokens to examine which ones are causally relevant in these tasks.

\begin{figure}[ht]
  \centering
  \includegraphics[width=\linewidth,page=5]{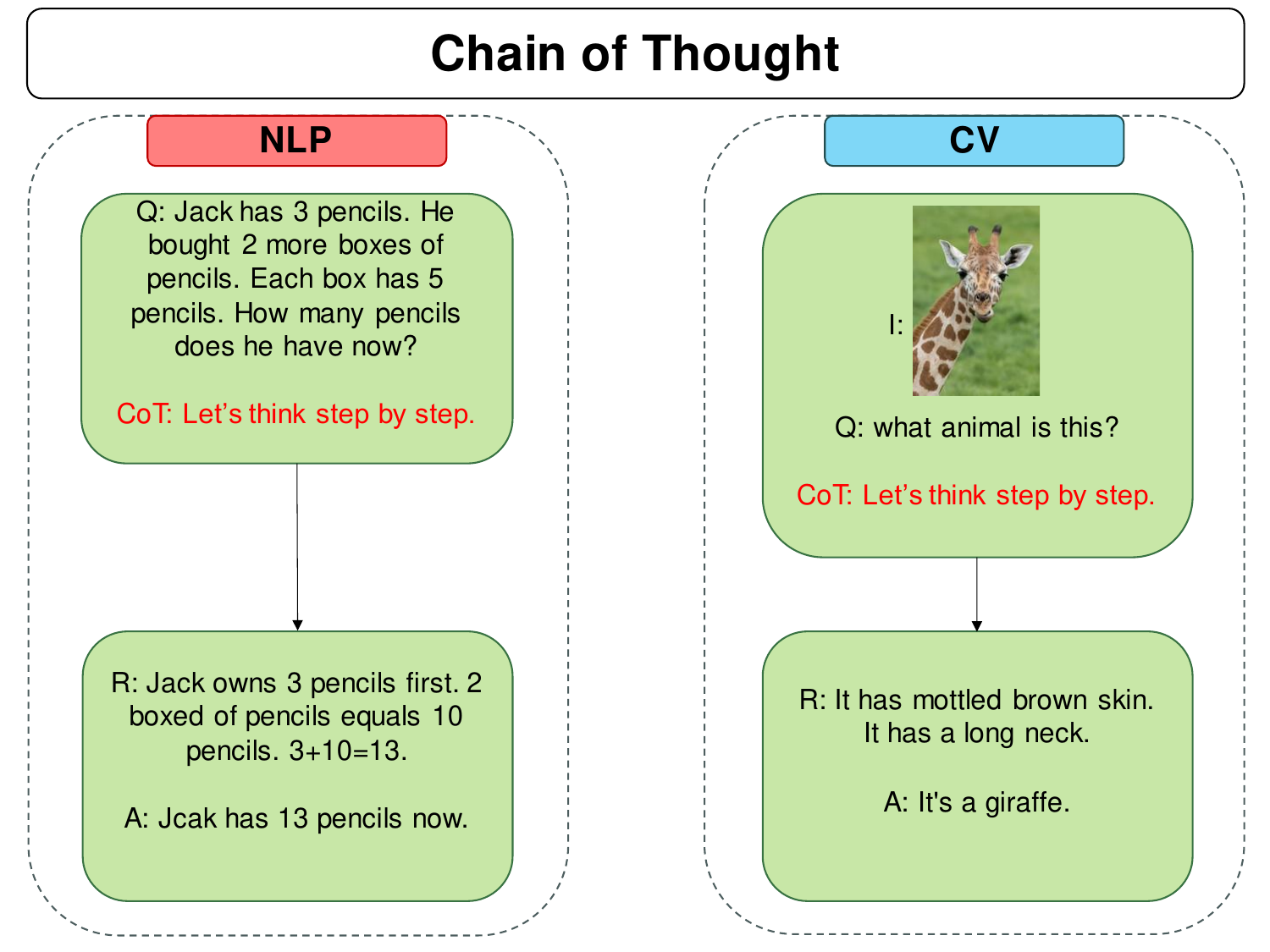}
  \caption{An overview of the CoT prompt tuning for interpretability, in both NLP and CV domains. ``Q'' in the figure refers to the question, ``R'' refers to reasoning and ``A'' refers to the answer. The image of a giraffe is from the work \cite{NEURIPS2022_a1859deb}.}
  \label{CoT}
\end{figure}

LLMs have shown impressive performance on complex reasoning by leveraging chain-of-thought (CoT)\cite{NEURIPS2022_9d560961} prompting to generate intermediate reasoning chains as the rationale to infer the answer \cite{zhang2023multimodal}. By integrating language and vision into a unified framework of CoT, interpretability of LMs can be facilitated as showed in Figure \ref{CoT}. Accordingly, Ge et al. \cite{ge2023chain} first successfully adapted chain-of-thought for prompt tuning that combines visual and textual embeddings in vision domain, obtaining a significant performance gain on many tasks. Moreover, VideoCOT \cite{himakunthala2023lets} has been proposed to enhance video reasoning while reducing the computational complexity of processing tremendous of frames on various video reasoning tasks. For image generation, leveraging CoT can utilize the power of DMs to generate a high-quality image, the intermediate space of which consists of a semantically-meaningful embedding from a pretrained CLIP embedder \cite{harvey2023visual}. Meanwhile, CoT can also control the foreground and background of generated images in a fine-grained manner to provide more interpretability \cite{lian2023llmgrounded}. In addition to the methods mentioned above, there is also a large amount of work about CoT focusing on this field \cite{zhang2023multimodal,chen2023measuring,mu2023embodiedgpt,rose2023visual}, which greatly promotes the visual interpretability of LMs.

\textbf{Discussion.} Exploring approaches to interpretability is an active and vital area in LMs study since humans long to safely utilize judgments and decisions from LMs. Therefore, comprehensive and trustworthy explanations for LMs' judgement and decision are necessary for human society.

\section{Conclusion}
Trustworthiness has been a topic of significant importance since LMs revolutionized various fields of deep learning with remarkable grades. More specifically, optimal trade-off between benefits and risks of using LMs is essential because humans cannot blindly trust LMs. To gain a thorough understanding of these risks, we have comprehensively reviewed current challenges of LMs in vision from the four major aspects. We logically elaborated these challenges based on their definition, countermeasures and discussion. In addition, this paper does not encompass all the content within this topic. For instance, the output of LMs is also limited by the timeliness of the data and perform worse in predicting future events beyond their training period. As a result, it is necessary to update the training dataset promptly to prevent time lag of information and enhance the generalization ability of LMs. An extremely comprehensive and complex issue in trustworthy LMs that is not included in our survey is how to make a model solve all the aforementioned problems simultaneously, which involves multi-criteria filtering of data and multi-task learning of models. This challenge lies at the core throughout the entire life-cycle of LMs because the deployment of LMs cannot ignore any of the above issues in practice, which is still an unexplored wilderness. In conclusion, we hope that this paper can provide a systematic review of past research and inspire more future research on trustworthy LMs in vision for the benefit of society.

\vspace{.3in} \noindent \textbf{Acknowledgements} \\
\noindent Not applicable.

\vspace{.3in} \noindent \textbf{Funding} \\
\noindent No funding was received to assist with the preparation of this manuscript.

\vspace{.3in} \noindent \textbf{Abbreviations} \\
\noindent LMs, large models; NLP, natural language processing; CV, computer vision; AI, artificial intelligence; LLMs, large language models; BERT, bidirectional encoder representation from transformers; GPT, generative pretrained transformer; CLIP, contrastive language-image pre-training; BLIP, bootstrapping language-image pre-training; DMs, diffusion models; NSFW, not safe for work; SOTA, state-of-the-art; CNN, convolution neural network; RNN, recurrent neural network; I2P, Inappropriate image prompts; RIATIG, reliable and imperceptible adversarial text-to-image generation with natural prompts; APBench, a unified benchmark for availability poisoning attacks and defenses; K-NN , k-nearest neighbor; De-Pois, an attack-agnostic approach for defense against poisoning attacks; BadEncoder, backdoor attacks to pre-trained encoders in self-supervised learning; ASR, attack success rate; FID, Frechet inception distance; YFCC, yahoo flickr creative commons 100M; MSCOCO, microsoft common objects in context; SVHN, street view house number; LLaVA, large language and vision assistant; PGD, projected gradient descent; TeCoA, a text-guided contrastive adversarial training; RLHF, reinforcement learning from human feedback; SFT, supervised fine-tuning; RM, reward model; PPO, proximal policy optimization; BLEU, bilingual evaluation understudy; ROUGE, recall-oriented understudy for gisting evaluation; METEOR, metric for evaluation of translation with explicit ordering; CIDEr, consensus-based image description evaluation; SPICE, semantic propositional image caption evaluation; CHAIR, caption hallucination assessment with image relevance; POPE, polling-based object probing evaluation; HaELM, hallucination evaluation based on large language models; LoRA, low-rank adaptation of large language models; GAVIE, GPT4-assisted visual instruction evaluation; MMHAL-BENCH, multimodal hallucination benchmark; MHalDetect, multimodal hallucination detection dataset; FDPO, fine-grained direct preference optimization; SCST, self-critical sequence training; LRV-Instruction, large-scale robust visual instruction; ObjMLM loss, object-masked language modeling loss; Grad-CAM, gradient-weighted class activation mapping; BagNet, deep bag-of-features models; ProtoPNet, prototypical part network; ProtoTree, neural prototype tree; CoT, chain-of-thought.

\vspace{.3in} \noindent \textbf{Availability of data and materials} \\
\noindent The datasets generated during and/or analyzed during the current study are available from the corresponding author upon reasonable request.

\vspace{.3in} \noindent \textbf{Competing Interests} \\
\noindent The authors declare no competing interests.

\vspace{.3in} \noindent \textbf{Authors’ Contributions} \\
\noindent Jun Liu proposed the idea for the article and revised the work. Ziyan Guo performed the literature search and drafted the work. Li Xu revised and improved the work. All authors read and approved the final manuscript.

\bibliographystyle{unsrt}
\bibliography{reference}

\end{document}